\documentclass[10pt]{article}
\usepackage[a4paper, total={5in, 7in}]{geometry}
\usepackage[usenames,dvipsnames,svgnames,table]{xcolor} 
\usepackage{authblk}

\usepackage{algorithm}
\usepackage{algorithmic}
\floatname{algorithm}{Algorithm}

\usepackage{longtable}
\usepackage{hhline}
\usepackage{float}
\newfloat{algorithm}{tb}{lop}

\usepackage{enumitem} 
\usepackage{rotfloat} 
\usepackage{graphicx}
\usepackage{epsfig}
\usepackage{amssymb, amsmath, amsthm}
\usepackage{tikz}
\usetikzlibrary{3d, calc}
\usepackage{verbatim}
\usepackage{hyperref}
\usepackage{natbib}
\usepackage{multirow}
\usepackage{subcaption} 
\usepackage{array}
\newcolumntype{L}{>{\centering\arraybackslash}m{0.1\linewidth}}

\newtheorem{theorem}{Theorem}[section]

\newtheorem{proposition}[theorem]{Proposition}

\newtheorem{definition}{Definition}

\newcommand{\calM}{\mathcal{M}}

\newcommand{\Frechet}{Fr\'{e}chet }

\newcommand{\SPD}{\mathcal{S}_{++}^n}
\newcommand{\CORR}{\mathcal{C}_{++}^n}
\newcommand{\DIAG}{\mathcal{D}_{++}^n}
\newcommand{\LTplus}{\mathcal{L}_{+}^n}
\newcommand{\LTones}{\mathcal{L}_1^n}
\newcommand{\LTzero}{\mathcal{L}_{0}^n}
\newcommand{\logTheta}{\log\circ \,\Theta}

\newcommand{\rmECM}{\textrm{ECM}}
\newcommand{\rmLEC}{\textrm{LEC}}

\begin{document}
	
\title{Scalable Geometric Learning with Correlation-Based Functional Brain Networks}
\author[1]{Kisung You}
\author[2]{Yelim Lee}
\author[2,3,4]{Hae-Jeong Park}

\affil[1]{Department of Mathematics, Baruch College, City University of New York, NY, USA.}
\affil[2]{Graduate School of Medical Science, Brain Korea 21 Project, Department of Nuclear Medicine, Psychiatry, Yonsei University College of Medicine, Seoul, Republic of Korea}
\affil[3]{Center for Systems and Translational Brain Science, Institute of Human Complexity and Systems Science, Yonsei University, Seoul, Republic of Korea}
\affil[4]{Department of Cognitive Science, Yonsei University, Seoul, Republic of Korea.}

\date{}

\maketitle
\begin{abstract}
The correlation matrix is a central representation of functional brain networks in neuroimaging. Traditional analyses often treat pairwise interactions independently in a Euclidean setting, overlooking the intrinsic geometry of correlation matrices. While earlier attempts have embraced the quotient geometry of the correlation manifold, they remain limited by computational inefficiency and numerical instability, particularly in high-dimensional contexts. This paper presents a novel geometric framework that employs diffeomorphic transformations to embed correlation matrices into a Euclidean space, preserving salient manifold properties and enabling large-scale analyses. The proposed method integrates with established learning algorithms - regression, dimensionality reduction, and clustering - and extends naturally to population-level inference of brain networks. Simulation studies demonstrate both improved computational speed and enhanced accuracy compared to conventional manifold-based approaches. Moreover, applications in real neuroimaging scenarios illustrate the framework’s utility, enhancing behavior score prediction, subject fingerprinting in resting-state fMRI, and hypothesis testing in electroencephalogram data. An open-source MATLAB toolbox is provided to facilitate broader adoption and advance the application of correlation geometry in functional brain network research.
\end{abstract}


\section{Introduction}\label{sec:intro}

A widely accepted view of the human brain is that it operates as a network formed by interactions among distributed regions \citep{park_2013_StructuralFunctionalBrain}. These interactions are often quantified using second-order statistics, including covariance, precision, and correlation matrices, which capture spontaneous fluctuations observed in resting-state functional magnetic resonance imaging (rs-fMRI) \citep{biswal_1995_FunctionalConnectivityMotor} or through electroencephalogram (EEG) and magnetoencephalogram (MEG) recordings \citep{brookes_2011_MeasuringFunctionalConnectivity, cohen_2014_AnalyzingNeuralTime}.

In most studies employing correlation matrices, interactions along individual edges are analyzed either independently of other edges \citep{dosenbach_2010_PredictionIndividualBrain, leonardi_2013_PrincipalComponentsFunctional, siman-tov_2017_EarlyAgeRelatedFunctional} or collectively, to identify sets of edges that interact synergistically \citep{leonardi_2013_PrincipalComponentsFunctional, park_2014_GraphIndependentComponent}. However, these approaches often fail to account for the intrinsic dependence structure among edges in a correlation matrix. A correlation matrix contains richer information as a whole than its individual pairwise correlations suggest, underscoring the need to treat it as a manifold-valued object with well-defined geometric properties.

Mathematically, the correlation matrix belongs to the class of symmetric positive-definite (SPD) matrices, whose collection constitutes a distinct geometric structure known as a Riemannian manifold. Recognizing the importance of adhering to the manifold's inherent geometry, an increasing number of studies have analyzed correlation matrices within the SPD manifold framework \citep{varoquaux_2010_DetectionBrainFunctionalConnectivity, ginestet_2017_HypothesisTestingNetwork, yamin_2019_ComparisonBrainConnectomes, abbas_2021_GeodesicDistanceOptimally}. A significant challenge in treating correlation matrices as SPD manifold objects arises because operations performed on these matrices often result in outputs that deviate from the form of a correlation matrix, necessitating post hoc normalization to enforce unit diagonal elements. In a prior study \citep{you_2021_RevisitingRiemannianGeometry}, we addressed this issue by iteratively normalizing matrices at each intermediate step. While effective in most scenarios, this heuristic approach lacks mathematical rigor and fails to guarantee exact solutions.

Unlike the SPD manifold, the space of correlation matrices, referred to as the elliptope \citep{tropp_2018_SimplicialFacesSet}, has received relatively limited attention. Only a few notable studies have explored this space, yet these efforts are hindered by either undesirable properties or a lack of efficient computational methods \citep{grubisic_2007_EfficientRankReduction, nielsen_2019_ClusteringHilbertProjective}. A promising alternative leverages the quotient geometry of the SPD manifold, induced by the affine-invariant Riemannian metric,  to represent the space of correlation matrices \citep{david_2019_RiemannianQuotientStructure, pennec_2006_IntrinsicStatisticsRiemannian, thanwerdas_2021_GeodesicQuotientAffineMetrics}.

Building upon these advancements, \citet{you_2022_GeometricLearningFunctional} incorporated the quotient geometry of the correlation manifold into well-known algorithms in machine learning and statistical inference, specifically for functional connectivity analysis. Despite its mathematical soundness and high performance, this approach faces critical challenges. These include computational inefficiency and numerical instability when applied to high-dimensional data, raising concerns about the robustness of the results. Addressing these limitations requires new methods to enable routine learning tasks at a practical scale.

Recently, \citet{thanwerdas_2022_TheoreticallyComputationallyConvenient} introduced novel geometric structures for the correlation manifold based on specialized transformations. These transformations preserve much of the geometric characteristics of the manifold while mapping correlation matrices to vectors by diffeomorphism, allowing the use of Euclidean geometry. This framework offers two key advantages: it facilitates the direct application of established algorithms from conventional learning paradigms and improves computational efficiency by confining expensive numerical operations to a one-time transformation.

The primary objective of the present study is to introduce these theoretical advancements to the neuroimaging community and demonstrate how this underutilized framework can enhance statistical learning with correlation-valued data in population-level functional connectivity analysis. The proposed approach achieves substantial computational speedups, thereby enabling correlation-based analyses for large-scale network studies, a significant limitation of previous methods, including our own.

This study is organized into three main sections. First, we revisit the foundational theory of Riemannian geometry and correlation manifolds. Next, we present the novel geometric structures and extend a suite of learning algorithms across multiple task categories. Finally, we evaluate the performance of these algorithms from computational and theoretical perspectives and apply the proposed pipeline to experimental data. To promote broader adoption, all algorithms have been implemented in a MATLAB toolbox (MathWorks, Inc., USA), which is freely available on a code-sharing platform for use by the neuroimaging community.


\section{Background}\label{sec:background}

\subsection{Basics of Riemannian manifolds}

A (smooth) \textit{manifold} is a topological space that locally resembles Euclidean space in the vicinity of each point \citep{lee_2012_IntroductionSmoothManifolds}. Formally, $\calM$ is defined as a $d$-dimensional manifold if every point $x \in \calM$ has a neighborhood that is homeomorphic to an open subset of $d$-dimensional Euclidean space. A defining feature of manifolds is that they are not vector spaces. This means that fundamental operations such as addition, subtraction, and scalar multiplication are not inherently defined. From a data analysis perspective, this characteristic presents a significant challenge, as conventional statistical learning methods rely on vector space structures and cannot be directly applied to manifold-valued data.

For example, consider a circle $S^1 \subset \mathbb{R}^2$ centered at the origin with a radius of one. In this case, the space of interest consists of points equidistant from the origin. Suppose we observe two data points at coordinates $(0,1)$ and $(0,-1)$, representing the north and south poles. A natural question arises: what is the mean of these two points? This concept is fundamental to clustering algorithms such as $k$-means \citep{macqueen_1967_MethodsClassificationAnalysis}, where clusters are defined based on proximity to a centroid. If we treat these points as vectors in Euclidean space, their mean is $(0,0) = \frac{1}{2}(0,1) + \frac{1}{2}(0,-1)$. However, this point does not lie on $S^1$. This example highlights the (potentially) nonlinear nature of manifolds and underscores the necessity of specialized mathematical tools for their analysis.

An introduction to manifolds typically begins with the notion of being locally Euclidean. This requires the concept of a tangent vector. Consider a smooth curve $\gamma: (-\epsilon, \epsilon) \rightarrow \calM$ for some $\epsilon > 0$, where the curve passes through a point $x$ at $t=0$, i.e., $\gamma(0) = x$. The derivative of $\gamma$ at $t=0$ is defined as a tangent vector at $x$, and the collection of all such vectors constitutes the tangent space $T_x \calM$. 

A manifold is called \textit{Riemannian} if it is equipped with a smoothly varying, positive-definite inner product $g_x: T_x \calM \times T_x \calM \rightarrow \mathbb{R}$, satisfying $g_x(v,v) > 0$ for all nonzero $v \in T_x \calM$. This metric enables the definition of fundamental concepts necessary for adapting statistical learning algorithms to manifold settings. For instance, the geodesic distance between two points on $\calM$ is defined as the length of the shortest curve connecting them, computed by integrating infinitesimal changes along the curve using the Riemannian metric.  For more details on Riemannian geometry and its computational applications, we refer readers to standard texts \citep{carmo_1992_RiemannianGeometry, absil_2008_OptimizationAlgorithmsMatrix, lee_2018_IntroductionRiemannianManifolds, boumal_2023_IntroductionOptimizationSmooth}.

\subsection{Geometry of SPD and CORR manifolds}

Brain functional connectivity is commonly represented using second-order statistics, such as covariance, correlation, or precision matrices. Mathematically, these matrices belong to the class of symmetric positive-definite (SPD) matrices, which are formally defined as follows:

\begin{definition} 
	$\SPD$ is the space of $(n \times n)$ symmetric positive-definite matrices: 
	\begin{equation*} 
		\SPD = \lbrace X \in \mathbb{R}^{n \times n} \mid X = X^\top,~ \lambda_{\min}(X) > 0 \rbrace, 
	\end{equation*} 
	where $\lambda_{\min}(\cdot)$ denotes the smallest eigenvalue of the matrix. 
\end{definition}

As a mathematical space, $\SPD$ has a dimension of $(n^2+n)/2$ and has attracted significant attention due to the frequent occurrence of such matrices in data analysis \citep{bhatia_2009_PositiveDefiniteMatrices}. Among the various geometric structures available, the affine-invariant Riemannian metric (AIRM) \citep{pennec_2006_IntrinsicStatisticsRiemannian} is one of the most prominent for $\SPD$, defining it as a Riemannian manifold. 

Under AIRM, the geodesic distance $d_{\SPD}(P, Q)$ between two SPD matrices $P, Q \in \SPD$ is expressed as: 
\begin{equation*} 
	d_{\SPD}^2(P, Q) = \| \log(P^{-1}Q) \|_F^2, 
\end{equation*} 
where $\|A \|_F = \sqrt{\text{Tr}(A^\top A)}$ is the Frobenius norm, and $\log(\cdot)$ denotes the matrix logarithm \citep{hall_2015_LieGroupsLie}. For any symmetric positive-definite matrix, the matrix logarithm is computed through its eigendecomposition, making it a well-defined and computationally feasible operation.

Our primary focus is on representing functional connectivity (FC) using correlation matrices, which constitute a specialized subset of SPD matrices. The space of correlation matrices, denoted as $\CORR$, is formally defined as follows:

\begin{definition} 
	$\CORR$ is the space of $(n \times n)$ symmetric positive-definite matrices with unit diagonal elements: 
	\begin{equation*} 
		\CORR = \lbrace X \in \mathbb{R}^{n \times n} \mid X \in \SPD,~ \text{diag}(X) = 1_n \rbrace, 
	\end{equation*} 
	where $\text{diag}(A)$ is a vector consisting of the diagonal elements of matrix $A$, and $1_n$ is the vector of length $n$ with all elements equal to 1. 
\end{definition}

\begin{figure}[ht]
	\centering
	\includegraphics[width=.5\linewidth]{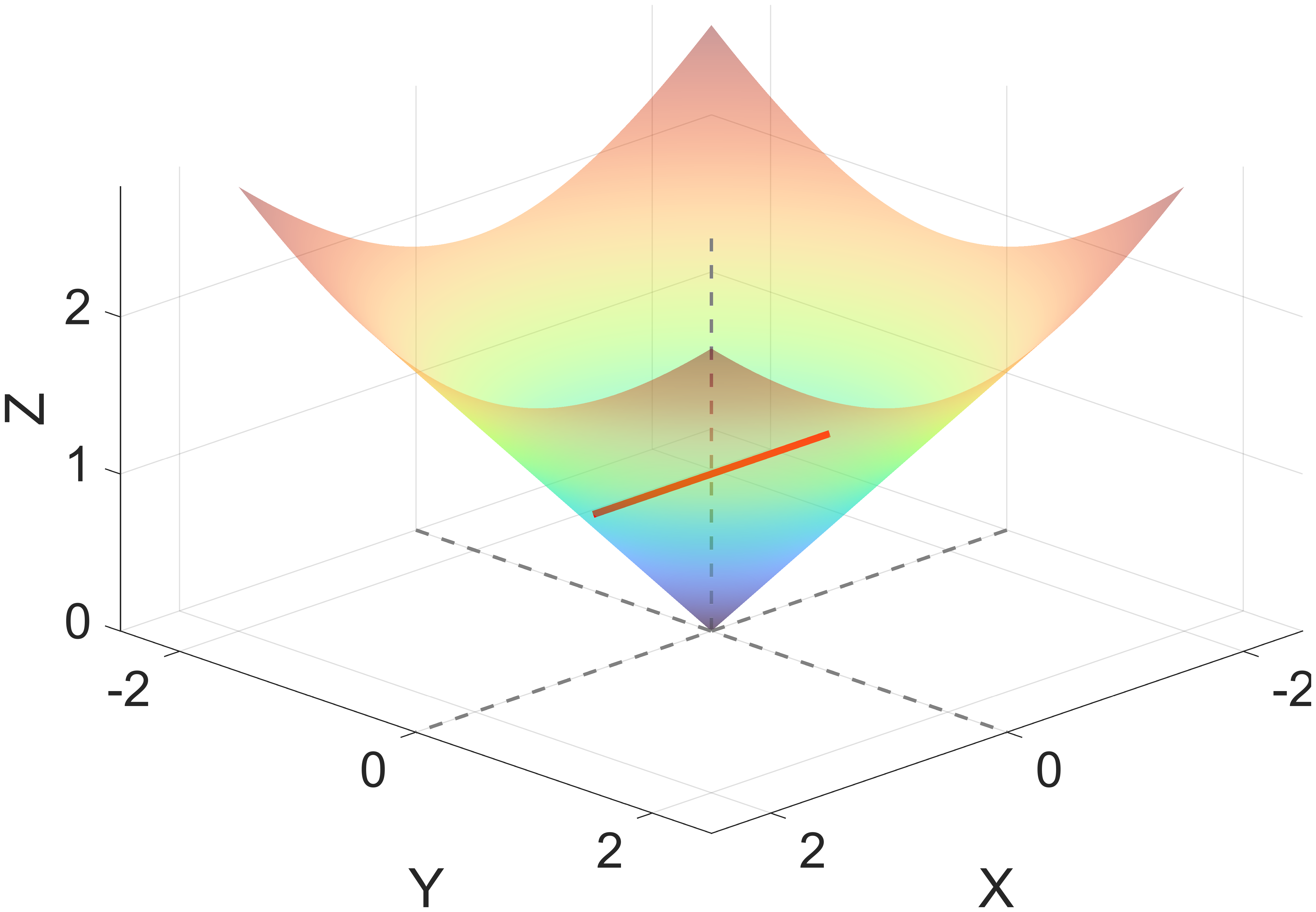}
	\caption{Visualization of the $2\times 2$ symmetric and positive-definite (SPD) manifold as the interior of the open upper cone in $\mathbb{R}^3$. The dashed gray lines indicate the coordinate axes, and the red line represents the correlation manifold, with its endpoints excluded, embedded within the SPD region.}
	\label{fig:VisualizeCone}
\end{figure}

This definition establishes that $\CORR$ is a strict subset of $\SPD$. For illustration, consider the simple case $n=2$, namely the collection of $2\times 2$ SPD and correlation matrices. A convenient way to visualize $\mathcal{S}_{++}^2$ is as the interior of the open upper cone in $\mathbb{R}^3$ \citep{bhatia_2009_PositiveDefiniteMatrices}. Let $C$ be a $2\times 2$ correlation matrix,
\begin{equation*}
	C = \begin{pmatrix}
		1 & r \\ r & 1
	\end{pmatrix}.
\end{equation*}
Every matrix in $\mathcal{C}_{++}^2$ has exactly one free parameter, the off-diagonal element $r$. Therefore, $\mathcal{C}_{++}^2$ corresponds to a one-dimensional manifold in $\mathbb{R}^3$, appearing as an open line segment, as shown in Figure~\ref{fig:VisualizeCone}. 

In more general settings, the space of correlation matrices can be endowed with a Riemannian manifold structure via the theory of quotient manifolds. In particular, \citet{david_2019_RiemannianQuotientStructure} introduced the quotient-affine metric (QAM), which inherits the affine-invariant Riemannian metric (AIRM) from 
$\SPD$ and thereby provides a Riemannian framework for correlation matrices.

Under the QAM, the geodesic distance between two correlation matrices $P, Q \in \CORR$ is defined as: 
\begin{equation*} 
	d_{\CORR}^2(P, Q) = \underset{D \in \DIAG}{\min}~ d_{\SPD}^2(P, DQD), 
\end{equation*} 
where $\DIAG$ represents the set of $(n \times n)$ diagonal matrices with strictly positive entries, and $d_{\SPD}$ denotes the geodesic distance on the SPD manifold under AIRM.

Unlike the direct computation of geodesic distance in $\SPD$ using AIRM, calculating the geodesic distance on $\CORR$ under QAM involves solving a nonlinear optimization problem. Each iteration of this process requires eigendecomposition to compute matrix square roots and logarithms, making the procedure computationally intensive. 

For further details on the AIRM and QAM geometries in the context of functional connectivity analysis, we direct readers to our earlier works \citep{you_2021_RevisitingRiemannianGeometry, you_2022_GeometricLearningFunctional}.


\begin{figure*}[ht]
	\centering
	\includegraphics[width=.9\linewidth]{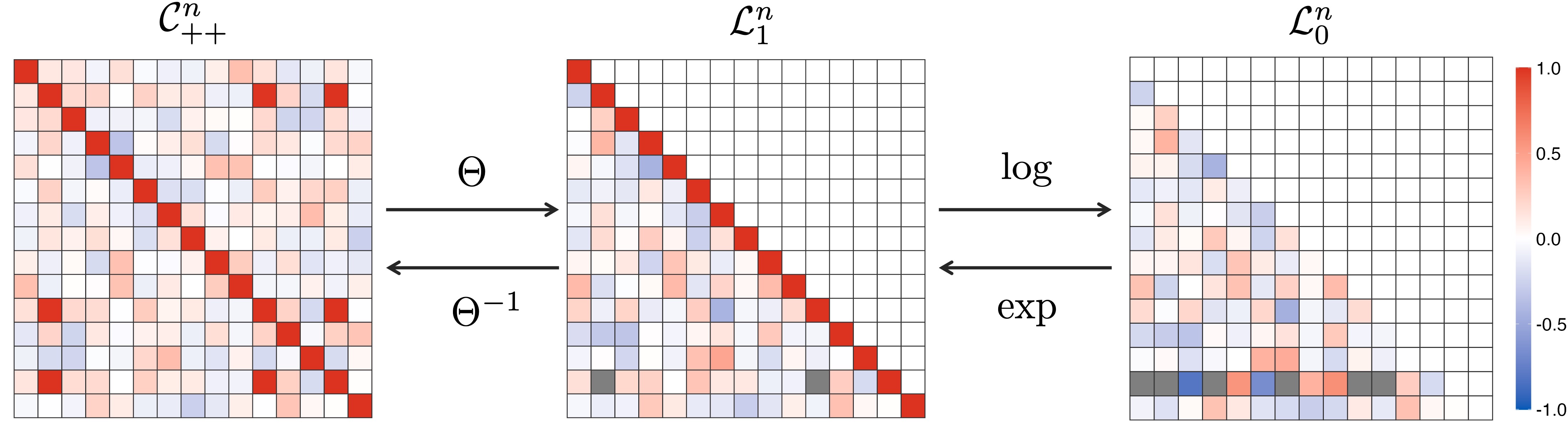}
	\caption{
		Diagram of the transformation process for ECM and LEC geometries. Applying the mapping $\Theta$ to a full-rank correlation matrix (left) results in a lower-triangular matrix with unit diagonals (middle). The subsequent application of the matrix logarithm to $\LTones$ produces strictly lower-triangular matrices with zero diagonals (right).}
	\label{fig:trfdiagram}
\end{figure*}

\section{New Geometries}\label{sec:geometries}

While the development of QAM geometry offers a promising framework for geometric learning on $\CORR$, it becomes computationally prohibitive as the number of functional connectivity matrices or the dimensionality of the regions of interest (ROIs) increases. In this section, we examine two alternative geometries for $\CORR$, recently proposed by \citet{thanwerdas_2022_TheoreticallyComputationallyConvenient}.

Before introducing these geometries, we establish the notations used throughout this section. The Cholesky decomposition is defined as the mapping $Chol: \SPD \rightarrow \LTplus$, where $\LTplus$ denotes the set of $(n \times n)$ lower-triangular matrices with positive diagonal entries. For any $\Sigma \in \SPD$, the Cholesky decomposition $Chol(\Sigma) = L$ ensures that $\Sigma = LL^\top$. The symbols $\LTzero$ and $\LTones$ represent the sets of lower-triangular matrices with zero diagonals and unit diagonals, respectively. Additionally, the operation $Diag(\cdot)$, when applied to a square matrix $A$, zeros out all off-diagonal elements, such that $Diag(A)_{i,j} = A_{i,j}$ if $i=j$ and $0$ otherwise. These notations will be integral in describing and analyzing the alternative geometries for $\CORR$.

The Euclidean-Cholesky metric (ECM) represents the first of these geometries, which transforms a correlation matrix into a lower-triangular matrix with unit diagonals. This transformation is defined as the mapping $\Theta: \CORR \rightarrow \LTones$, such that for any $C \in \CORR$, 

\begin{equation*} 
	\Theta(C) = Diag(Chol(C))^{-1} \cdot Chol(C). 
\end{equation*} This mapping ensures that the resulting lower-triangular matrix belongs to $\LTones$, facilitating the application of Euclidean geometry in this transformed space.

The map $\Theta$ is smooth, allowing the use of the vector space structure of $\LTones$ to define a pullback metric through $\Theta$, incorporating the logarithmic transformation of the diagonal elements in $\LTones$. Under the ECM geometry, the distance between two correlation matrices $C_1, C_2 \in \CORR$ is defined as  
\begin{equation}\label{eq:defDistECM}
	d_{\rmECM}(C_1, C_2) = \| \Theta(C_1) - \Theta(C_2) \|_F,
\end{equation} where $\|\cdot \|_F$ denotes the standard Frobenius norm. The unique geodesic curve $\gamma:[0,1] \rightarrow \CORR$ connecting the two points $C_1$ and $C_2$ is expressed as \begin{equation*}
	\gamma_{\rmECM}(t) = \Theta^{-1} \left( (1-t) \cdot  \Theta(C_1) + t \cdot  \Theta(C_2) \right),
\end{equation*} with $\gamma_{\rmECM}(0) = C_1$ and $\gamma_{\rmECM}(1) = C_2$. The inverse mapping $\Theta^{-1}: \LTones \rightarrow \CORR$ for any $L \in \LTones$ is explicitly available as the following:
\begin{equation*}
	\Theta^{-1}(L) = Diag(LL^\top)^{-1/2} \cdot LL^\top \cdot Diag(LL^\top)^{-1/2},
\end{equation*}
ensuring that $\Theta^{-1} \circ \Theta(C) = C$ for all $C \in \CORR$.

Building on ECM, the Log-Euclidean Cholesky metric (LEC) introduces a different vector space structure on $\CORR$ by applying the matrix logarithm to $\Theta$. Recall that the logarithm of a square matrix $Z$ is defined through the power series:
\begin{equation*}
	\log(Z) = \sum_{k=1}^\infty \frac{(-1)^{k-1}}{k} (Z - I_n)^k,
\end{equation*}
where $I_n$ is the identity matrix of size $n \times n$. This series converges for matrices $Z$ whose eigenvalues lie in the positive real half-plane, making it a suitable operation for matrices in $\SPD$.

Given $Z \in \LTones$, it follows that $Z - I_n \in \LTzero$ because $Z$ has unit diagonals. Consequently, $(Z - I_n)^k$ remains strictly lower-triangular for $k < n$ and becomes the zero matrix for $k \geq n$. This property allows the matrix logarithm to serve as a smooth mapping from $\LTones$ to $\LTzero$:
\begin{equation*}
	\log (Z) = \sum_{k=1}^{n-1} \frac{(-1)^{k-1}}{k} (Z - I_n)^k,
\end{equation*}
which involves only a finite number of matrix powers. The LEC defines the composite mapping $\logTheta: \CORR \rightarrow \LTzero$, which acts as a diffeomorphism and equips $\CORR$ with the pullback metric of the standard Euclidean inner product. The distance between two points $C_1, C_2 \in \CORR$ under the LEC geometry is given by:
\begin{equation}\label{eq:defDistLEC}
	d_{\rmLEC}(C_1, C_2) = \|\logTheta(C_1) - \logTheta(C_2)\|_F,
\end{equation}
where $\|\cdot\|_F$ denotes the Frobenius norm. Similar to the ECM framework, the geodesic curve $\gamma:[0,1] \rightarrow \CORR$ under LEC geometry, connecting $C_1$ and $C_2$, is determined as:
\begin{equation*}
	\gamma_{\rmLEC}(t) = 
	(\logTheta)^{-1}
	\big(
	(1-t)\cdot \logTheta(C_1) + t\cdot \logTheta(C_2)
	\big),
\end{equation*}
where $(\logTheta)^{-1}$ is the composite inverse of the matrix exponential and $\Theta$, explicitly defined as $(\logTheta)^{-1} = \Theta^{-1} \circ \exp$. These transformations are illustrated in Figure \ref{fig:trfdiagram}. For a comprehensive discussion of the geometric operations associated with these two geometries, refer to \citet{thanwerdas_2022_TheoreticallyComputationallyConvenient}.

We conclude this section by emphasizing the advantages of adopting the geometries described above. First, these geometries exhibit the characteristics of standard Euclidean space via diffeomorphic transformations, leading to zero curvature. This property enables the straightforward application of interpolation, extrapolation, and the computation of unique centroids due to the homogeneous space property \citep{afsari_2011_Riemannian$L_p$Center}. 

Second, these geometries offer significant computational benefits. Both start with the Cholesky decomposition, which has a computational complexity of $O(n^3)$ \citep{demmel_1997_AppliedNumericalLinear}. In the case of the LEC geometry, the $\logTheta$ mapping requires an additional matrix logarithm step, with complexity $O(n^\omega)$ for $\omega \in (2, 2.376)$ \citep{demmel_fast_2007}. Once the transformation is performed, subsequent computations follow standard multivariate analysis routines in the Euclidean space. In contrast, QAM geometry necessitates solving an optimization problem even for basic distance computations, making it significantly less efficient.

\begin{figure}[ht!]
	\centering
	\includegraphics[width=.5\linewidth]{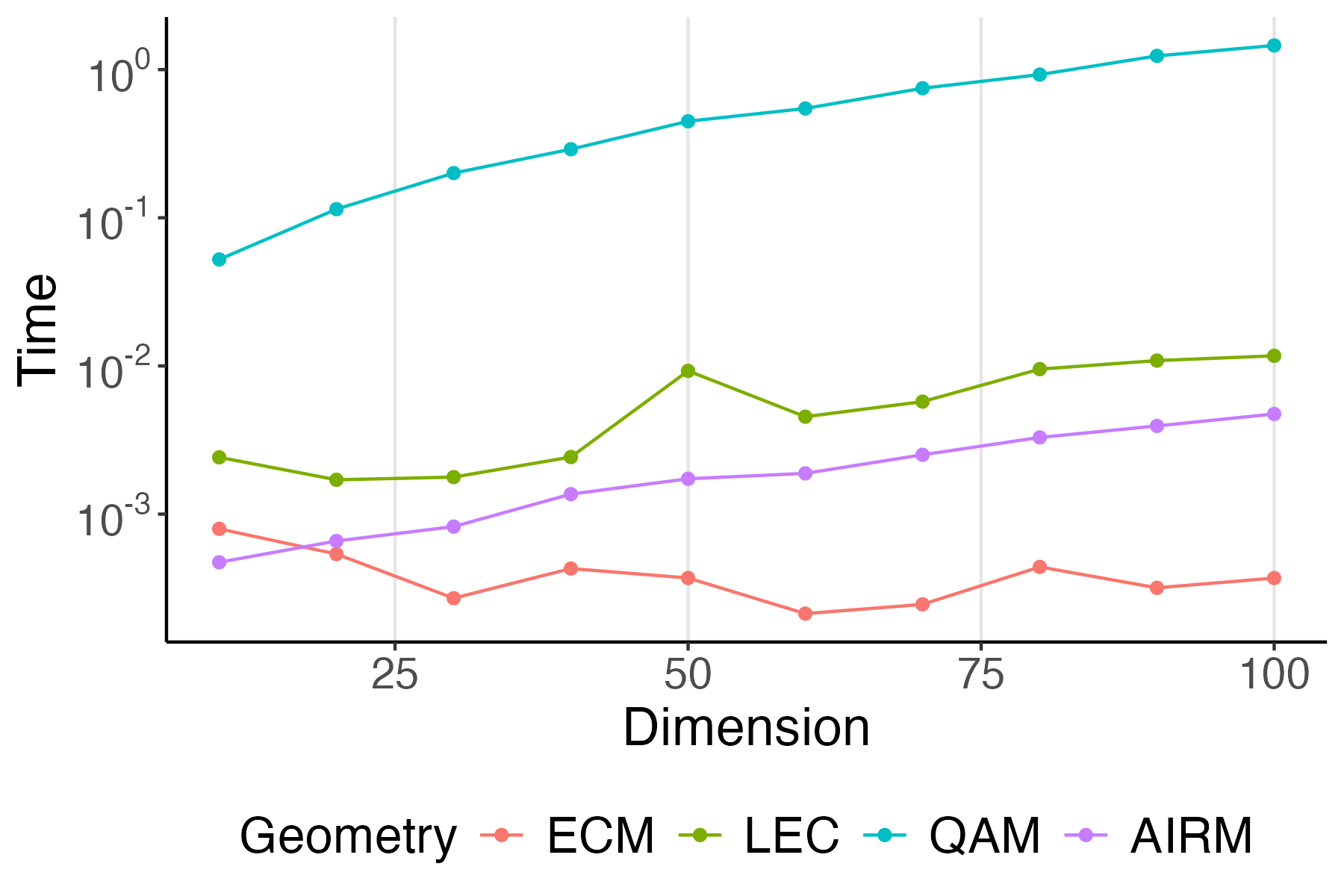}
	\caption{Comparison of average wall-clock runtime over 50 trials for computing the distance between perturbed versions of two model correlation matrices, $C_1$ and $C_2$, at varying dimensions $n=10, 20, \ldots, 100$. The $y$-axis represents the average runtime in seconds, displayed on a base-10 logarithmic scale.}
	\label{fig:geometry-time4dist}
\end{figure}

This distinction is illustrated in the following example, where we compute the distance between two correlation matrices $C_1$ and $C_2 \in \CORR$. Here, $C_1$ is the identity matrix, and $C_2$ is derived from an AR(1) process with $C_2(i,j) = \rho^{|i-j|}$ for $\rho=0.8$. Figure \ref{fig:geometry-time4dist} summarizes the average runtime over 50 trials for computing distances between perturbed versions of $C_1$ and $C_2$ across varying dimensions $n = 10, 20, \dots, 100$. The ECM and LEC geometries demonstrate remarkable computational efficiency, outperforming QAM geometry by several orders of magnitude.

\section{Methods}\label{sec:methods}

This section introduces several categories of algorithms tailored for analyzing populations of functional connectivity matrices, where each connectivity matrix is treated as an element of $\CORR$. Throughout this section, the symbol $d$ represents a general distance metric, which may correspond to distances defined by the ECM or LEC geometries, as described in Equations \eqref{eq:defDistECM} and \eqref{eq:defDistLEC}. Whenever these specific metrics are utilized, they will be explicitly indicated.

\begin{table*}[ht]
	\centering
	\caption{Summary of learning algorithm categories for population-level inference using correlation-based functional connectivity. The middle column provides the MATLAB function names included in the \textsf{CORRbox} package.}
	\label{tab:algorithms}
	
	\begin{tabular}{|c|c|l|}
		\hline
		Category                                                                      & Function    & \multicolumn{1}{c|}{Description} \\ \hline\hline 
		\multirow{2}{*}{\begin{tabular}[c]{@{}c@{}}Exploratory\\ Analysis\end{tabular}} & \textbf{corr\_mean.m}    
		& compute the \Frechet mean and variation \\ \cline{2-3}
		& \textbf{corr\_median.m} & compute the \Frechet median and variation \\ \hline  
		\multirow{3}{*}{Regression} & \textbf{corr\_gpreg.m}      & Gaussian process regression   \\ \cline{2-3} 
		& \textbf{corr\_kernreg.m} & kernel regression \\ \cline{2-3}
		& \textbf{corr\_svmreg.m} & support vector regression \\ \hline
		\multirow{5}{*}{\begin{tabular}[c]{@{}c@{}}Dimensionality\\ Reduction\end{tabular}} 
		& \textbf{corr\_ae.m} & shallow autoencoder \\ \cline{2-3}
		& \textbf{corr\_cmds.m} & classical multidimensional scaling \\ \cline{2-3}
		& \textbf{corr\_mmds.m} & metric multidimensional scaling  \\ \cline{2-3}
		& \textbf{corr\_pga.m} & principal geodesic analysis \\ \cline{2-3}
		& \textbf{corr\_tsne.m} & $t$-stochastic neighbor embedding \\ \hline 
		\multirow{5}{*}{\begin{tabular}[c]{@{}c@{}}Cluster\\ Analysis\end{tabular}} 
		& \textbf{corr\_kmeans.m} & $k$-means clustering \\ \cline{2-3}
		& \textbf{corr\_kmedoids.m} & $k$-medoids clustering\\ \cline{2-3}
		& \textbf{corr\_specc.m} & spectral clustering  \\ \cline{2-3}
		& \textbf{corr\_silhouette.m} & cluster validity index of Silhouette score\\ \cline{2-3}
		& \textbf{corr\_CH.m} & cluster validity index of Calinski and Harabasz \\ \hline 
		\multirow{4}{*}{\begin{tabular}[c]{@{}c@{}}Hypothesis\\ Testing\end{tabular}}
		& \textbf{corr\_test2bg.m} & two-sample test via Biswas-Ghosh method \\ \cline{2-3}
		& \textbf{corr\_test2energy.m} & two-sample test with the energy distance \\ \cline{2-3}
		& \textbf{corr\_test2mmd.m} & two-sample test via maximum mean discrepancy \\ \cline{2-3}
		& \textbf{corr\_test2wass.m} & two-sample test with the Wasserstein distance \\ \hline 
	\end{tabular}
\end{table*}

\subsection{Exploratory analysis}

Consider a random sample of functional connectivity representations $\lbrace \Sigma_i \rbrace_{i=1}^m \subset \CORR$. The initial step in data analysis often involves examining the sample’s summary statistics, such as its centroid and dispersion. These are referred to as generalized \Frechet means or $L_p$ centers of mass in the context of manifold-valued data analysis \citep{afsari_2011_Riemannian$L_p$Center}. 

The $L_p$ center of mass is defined as the minimizer of the functional:
\begin{equation}\label{eq:frechet-general}
	F_p(\Sigma) = \frac{1}{m} \sum_{i=1}^m d^p(\Sigma, \Sigma_i),
\end{equation}
where $p \geq 1$. For $p=2$, the minimizer is known as the \Frechet mean (\textbf{corr\_mean.m}), which generalizes the notion of the mean to general metric spaces. For $p=1$, the minimizer of $F_1(\Sigma)$ is called the \Frechet or geometric median (\textbf{corr\_median.m}), a robust alternative to the \Frechet mean \citep{fletcher_2009_GeometricMedianRiemannian}. The \Frechet variation, which measures dispersion, generalizes the concepts of variance and mean absolute deviation for $p=2$ and $p=1$, respectively. Denoting the minimizer of Equation \eqref{eq:frechet-general} as $\hat{\Sigma}$, the \Frechet variation is given by $F_p(\hat{\Sigma})$, representing the value of the functional at its minimum.

The ECM and LEC geometries guarantee the existence of a unique \Frechet mean. However, the uniqueness of the \Frechet median requires the assumption that the images of the mappings are not collinear, which is rarely violated in practical data analysis.

\subsection{Regression on scalars}

Can individual functional connectivity predict phenotypic traits? This question falls under the domain of regression analysis, which investigates the relationship between individuals' brain networks and variables of interest. Given a set of correlation-based connectivity matrices as independent variables and scalar phenotypes as dependent variables, the data pairs $(\Sigma_i, Y_i)_{i=1}^m \subset \CORR \times \mathbb{R}$ are analyzed to identify a function $f:\CORR \rightarrow \mathbb{R}$ that satisfies
\begin{equation*} 
	Y_i = f(\Sigma_i) + \epsilon_i, 
\end{equation*}
where $\epsilon_i$ is an additive error term. The function $f$ can be estimated under specific assumptions regarding the functional form and error distribution.

We explore three nonlinear regression models within the framework of kernel methods \citep{scholkopf_2002_LearningKernelsSupport}: (1) Gaussian process regression (\textbf{corr\_gpreg.m}), (2) kernel regression (\textbf{corr\_kernreg.m}), and (3) support vector regression (\textbf{corr\_svmreg.m}). Two main reasons motivate the inclusion of kernel-based approaches in this context.

First, while linear models are simple and interpretable, they often lack the flexibility required for capturing the inherent nonlinear relationships in correlation matrices. Although transformations such as $\Theta$ or $\logTheta$ could be applied to linear models, they still fail to improve interpretability due to the nonlinear nature of such transformations. Additionally, linear models may underperform when modeling the complexities of brain connectivity.

Second, kernel methods built on ECM and LEC geometries offer theoretical and practical advantages. Kernel methods replace inner products in high-dimensional feature spaces with kernel functions, leveraging the Gram matrix $K \in \mathbb{R}^{m \times m}$, where $K_{i,j} = k(x_i, x_j)$ for some kernel function $k(\cdot, \cdot)$. For consistent and generalizable performance, the Gram matrix must be positive semi-definite \citep{mohri_2012_FoundationsMachineLearning}. A positive-definite kernel is a continuous function $k: \mathcal{X} \times \mathcal{X} \rightarrow \mathbb{R}$ satisfying 
\begin{equation*}
	\sum_{i=1}^m \sum_{j=1}^m c_i c_j k(x_i, x_j) \geq 0
\end{equation*}
for any $x_1, \ldots, x_m \in \mathcal{X}$ and $c_1, \ldots, c_m \in \mathbb{R}$. A commonly used positive-definite kernel is the squared exponential kernel, which is valid under both ECM and LEC geometries.

The following proposition establishes that the squared exponential kernel is positive-definite.

\begin{proposition}\label{thm:pdkernel}
	For $C_i, C_j \in \CORR$ and a non-negative constant $\theta \geq 0$, the squared exponential kernel 
	\begin{equation}\label{eq:pdkernel}
		k(C_i, C_j) = \exp \left(- \theta \cdot  {d_*^2(C_i,C_j)}\right)
	\end{equation}
	is a positive-definite kernel when $d_* = d_{\rmECM}$ or $d_{\rmLEC}$.
\end{proposition}

\begin{proof}
	The property of the squared exponential kernel on topological spaces has been extensively studied in the literature \citep{berg_1984_HarmonicAnalysisSemigroups, klebanov_2005_NdistancesTheirApplications, feragen_2015_GeodesicExponentialKernels, jayasumana_2015_KernelMethodsRiemannian}. For completeness, we outline the adapted proof below.
	
	Without loss of generality, denote the diffeomorphisms associated with the ECM and LEC geometries as $\phi$. Thus, the distance between two correlation matrices $C_1$ and $C_2$ can be expressed as $d_*(C_1, C_2) = \|\phi(C_1) - \phi(C_2)\|_F$. Define the mapping $\psi : \CORR \times \CORR \rightarrow \mathbb{R}$ as $\psi(C_1, C_2) = d_*^2(C_1, C_2)$. In kernel methods, a mapping is called a conditionally negative definite (CND) kernel if it satisfies:
	\begin{enumerate}
		\item $\psi(C, C) = 0$ for all $C \in \CORR$,
		\item $\psi(C_1, C_2) = \psi(C_2, C_1)$ for all $C_1, C_2 \in \CORR$, and
		\item For any integer $m > 0$, the inequality 
		\begin{equation}\label{eq:cnd}
			\sum_{i=1}^m \sum_{j=1}^m c_i c_j \psi(C_i, C_j) \leq 0
		\end{equation}
		holds for any $\lbrace C_i \rbrace_{i=1}^m \subset \CORR$ and constants $\lbrace c_i \rbrace_{i=1}^m \subset \mathbb{R}$ such that $\sum_{i=1}^m c_i = 0$.
	\end{enumerate}
	
	The first two properties are straightforward, as $\psi$ is defined using the Euclidean distance between transformed correlation matrices through diffeomorphisms. Let the left-hand side of Equation \eqref{eq:cnd} be denoted as $(\triangle)$:
	\begin{align*}
		(\triangle) &= \sum_{i=1}^m \sum_{j=1}^m c_i c_j \|\phi(C_i) - \phi(C_j)\|_F^2 \\ 
		&= \sum_{i=1}^m \sum_{j=1}^m c_i c_j \left( \|\phi(C_i)\|_F^2 + \|\phi(C_j)\|_F^2 - 2\langle \phi(C_i), \phi(C_j) \rangle \right) \\
		&= \sum_{i=1}^m \sum_{j=1}^m c_i c_j \|\phi(C_i)\|_F^2 + \sum_{i=1}^m \sum_{j=1}^m c_i c_j \|\phi(C_j)\|_F^2 \\
		&\quad - 2 \sum_{i=1}^m \sum_{j=1}^m c_i c_j \langle \phi(C_i), \phi(C_j) \rangle \\
		&= -2 \left\langle \sum_{i=1}^m c_i \phi(C_i), \sum_{j=1}^m c_j \phi(C_j) \right\rangle \\ 
		&= -2 \left\| \sum_{i=1}^m c_i \phi(C_i) \right\|_F^2 \leq 0,
	\end{align*}
	where the fourth equality arises from the constraint $\sum_{i=1}^m c_i = 0$, causing the first two terms to vanish. Thus, $\psi(C_1, C_2) = d_*^2(C_1, C_2)$ is indeed a CND kernel.
	
	By Theorem 4 of \citet{schoenberg_1938_MetricSpacesCompletely}, the kernel of a form $k(C_1, C_2) = e^{-\theta \psi(C_1, C_2)}$ is positive-definite for all $\theta \geq 0$ if and only if $\psi$ is a CND kernel. Consequently, the squared exponential kernel in Equation \eqref{eq:pdkernel} is a positive-definite kernel on $\CORR$.
\end{proof}

It is important to note that the distance in quotient geometry does not inherently guarantee positive definiteness of the induced kernels due to the intrinsic curvature of $\CORR$. Another point of emphasis is that the three regression algorithms discussed, Gaussian process regression, kernel regression, and support vector regression rely on several hyperparameters. To optimize these hyperparameters, we utilized 5-fold cross-validation, ensuring robust fine-tuning for generalization and performance.

\subsection{Dimensionality reduction}

Visual examination of data distribution is an invaluable step in analyzing complex datasets, as it provides intuitive insights into patterns and relationships that might be obscured in high-dimensional representations. The field of dimensionality reduction addresses this challenge by focusing on techniques to represent high-dimensional data in low-dimensional spaces that are more interpretable by humans \citep{engel_2012_SurveyDimensionReduction, you_2022_RdimtoolsPackageDimension}. From a theoretical standpoint, dimensionality reduction involves finding a mapping $f$ from the original data space onto $\mathbb{R}^d$, where $d$ is typically chosen as 2 or 3 to facilitate visualization. This mapping can be defined either explicitly, where the transformation is mathematically described, or implicitly, where the relationship is inferred through computational algorithms.

In the context of our work, we leverage the Euclideanization properties of the proposed geometries, ECM and LEC, to incorporate several dimensionality reduction algorithms into our analysis toolkit without developing dedicated algorithms on $\CORR$. A list of algorithms capable of defining an explicit form of mapping from $\CORR$ to $\mathbb{R}^d$ includes principal geodesic analysis (\textbf{corr\_pga.m}) \citep{fletcher_2004_PrincipalGeodesicAnalysis} and shallow autoencoders (\textbf{corr\_ae.m}) \citep{baldi_1989_NeuralNetworksPrincipal}. On the other hand, methods such as classical and metric multidimensional scaling (\textbf{corr\_cmds.m}, \textbf{corr\_mmds.m}) \citep{borg_1997_ModernMultidimensionalScaling}, and $t$-stochastic neighbor embedding (\textbf{corr\_tsne.m}) \citep{vandermaaten_2008_VisualizingDataUsing} belong to a category of algorithms that do not provide explicit mappings.

It is important to emphasize that except for principal geodesic analysis, the dimensionality reduction methods integrated into our framework are nonlinear. This nonlinearity is crucial for capturing and representing the low-dimensional structures embedded within the correlation manifold. These methods are particularly advantageous for uncovering complex relationships in data, as they adapt to the curvature and geometry of $\CORR$, providing a more faithful representation of the intrinsic structure of the dataset.

\subsection{Cluster analysis}

Cluster analysis focuses on uncovering the inherent subgroup structure within data when the true labels of the data points are unknown \citep{kaufman_2005_FindingGroupsData}. This type of unsupervised learning is particularly valuable in exploratory data analysis, where prior knowledge about the underlying groups is limited. In this context, we emphasize partitional clustering algorithms, including geometric adaptations of $k$-means (\textbf{corr\_kmeans.m}) \citep{macqueen_1967_MethodsClassificationAnalysis}, $k$-medoids (\textbf{corr\_kmedoids.m}) \citep{kaufman_1990_PartitioningMedoidsProgram}, and spectral clustering (\textbf{corr\_specc.m}) \citep{vonluxburg_2007_TutorialSpectralClustering}, all of which are designed to identify a predetermined number of clusters within a set of correlation matrices.

In partitional clustering, the primary objective is to minimize within-cluster variation while maximizing between-cluster separation. Algorithms such as $k$-means and $k$-medoids operate by iteratively refining the assignment of data points to clusters based on a predefined distance metric. In our case, the Euclideanized metrics derived from ECM and LEC geometries are employed. The $k$-means algorithm uses centroids to represent each cluster, which may be unsuitable for non-Euclidean spaces. However, the geometric adaptations in \textbf{corr\_kmeans.m} ensure compatibility with $\CORR$. On the other hand, the $k$-medoids algorithm selects actual data points as cluster representatives, providing a robust alternative, especially when data distributions are non-convex or include outliers. Spectral clustering further enhances this toolkit by leveraging eigenvalue decomposition on an affinity matrix constructed from pairwise distances, allowing for flexible and effective identification of nonlinearly separable clusters.

In our context, these clustering methods facilitate the discovery of cohesive yet distinct subpopulations of functional connectivity patterns. This is particularly valuable in neuroimaging studies, where the data often exhibits high heterogeneity. For instance, in research on autism spectrum disorder, identifying meaningful subgroups can provide insights into the disorder’s variability across individuals and may even guide personalized therapeutic approaches \citep{eaves_1994_SubtypesAutismCluster}.

To evaluate clustering performance, we employ two metrics: the Silhouette score (\textbf{corr\_silhouette.m}) \citep{rousseeuw_1987_RobustRegressionOutlier} and the Calinski-Harabasz (CH) index (\textbf{corr\_CH.m}) \citep{calinski_1974_DendriteMethodCluster}. The Silhouette score measures the degree of cohesion and separation within clusters, with higher values indicating well-defined and distinct clusters. It is computed as the difference between the average intra-cluster distance and the smallest average inter-cluster distance, normalized by the maximum of the two. This score provides an interpretable measure of how similar each data point is to its assigned cluster relative to other clusters.

The CH index evaluates clustering quality by comparing the dispersion of data points within clusters to the dispersion between clusters. Specifically, it computes the ratio of between-cluster dispersion to within-cluster dispersion, scaled by the number of clusters and the total number of data points. Higher CH values indicate better-defined cluster structures, making it an effective tool for selecting the optimal number of clusters.

Given the limited \textit{a priori} knowledge of subgroup characterization in data-driven studies, these indices are critical for assessing clustering quality and determining the most appropriate or plausible number of clusters. They provide a quantitative basis for validating clustering outcomes, enabling researchers to interpret results with greater confidence.

\subsection{Hypothesis testing}

The final set of algorithms pertains to two-sample hypothesis testing, an essential framework for comparing two groups of functional connectivity matrices to identify statistically significant differences. Consider two sets of correlation matrices, $C_1^{(1)}, \ldots, C_{m_1}^{(1)}$ and $C_1^{(2)}, \ldots, C_{m_2}^{(2)}$, sampled from underlying probability distributions $\mathbb{P}^{(1)}$ and $\mathbb{P}^{(2)}$, respectively. While many two-sample tests focus on differences in means, variances, or other summary statistics, our emphasis lies on testing the equality of entire distributions by formulating the null hypothesis as $H_0: \mathbb{P}^{(1)} = \mathbb{P}^{(2)}$. This type of test is particularly relevant for studies involving functional connectivity matrices, where the data are naturally grouped based on population characteristics such as disease status, cognitive phenotype, or experimental condition.

Central to this framework is the concept of measuring dissimilarity $\mathcal{D}$ between two probability distributions. This defines a class of two-sample testing algorithms that quantify the extent to which $\mathbb{P}^{(1)}$ differs from $\mathbb{P}^{(2)}$ \citep{ramdas_2017_WassersteinTwoSampleTesting}. Commonly used measures in the context of correlation matrices under ECM and LEC geometries include:

\begin{itemize}
	\item Maximum mean discrepancy (\textbf{corr\_test2mmd.m}) measures differences between distributions in a reproducing kernel Hilbert space, leveraging kernel-based representations of data.
	\item Wasserstein distance (\textbf{corr\_test2wass.m}) captures the minimal cost of transforming one distribution into another, often referred to as the ``earth mover’s distance.''
	\item Energy distance (\textbf{corr\_test2energy.m}) computes pairwise distances between all samples, focusing on differences in inter-point relationships.
\end{itemize}

These measures share a crucial property: $\mathcal{D} \geq 0$, where equality implies that $\mathbb{P}^{(1)}$ and $\mathbb{P}^{(2)}$ are indistinguishable under the specific discrepancy measure. Viewing the two sets of observations as empirical measures:
\begin{equation*} 
	\mathbb{P}^{(1)} = \frac{1}{m_1} \sum_{i=1}^{m_1} \delta_{C_i^{(1)}} \quad \text{and} \quad \mathbb{P}^{(2)} = \frac{1}{m_2} \sum_{j=1}^{m_2} \delta_{C_j^{(2)}}, 
\end{equation*} 
where $\delta$ represents a Dirac mass, the null hypothesis is equivalently tested by checking whether $\mathcal{D}(\mathbb{P}^{(1)}, \mathbb{P}^{(2)}) = 0$.

An alternative perspective is provided by inter-point distance-based methods. Testing the equality of distributions can also be formulated by analyzing the distributions of distances $d(C^{(1)}, \tilde{C}^{(1)})$, $d(C^{(2)}, \tilde{C}^{(2)})$, and $d(C^{(1)}, C^{(2)})$, where $\tilde{X}$ represents an independent sample identically distributed as $X$. This approach underpins the Biswas-Ghosh test (\textbf{corr\_test2bg.m}) \citep{biswas_2014_NonparametricTwosampleTest}.

Despite their theoretical soundness, these four tests face practical challenges. The limiting distributions of their test statistics are either unknown or are only known under restrictive assumptions, limiting their direct application in real-world scenarios. To address this, we adopt a permutation testing framework, a resampling-based approach that establishes a threshold for the test statistic by permuting class labels \citep{pitman_1937_SignificanceTestsWhich}. This method provides robust control over Type I error rates without requiring strong parametric assumptions.

We outline a generic pipeline for the resampling procedure, applicable to all four tests introduced in this paper. Let $\mathcal{C}_i = \lbrace C_1^{(i)}, \ldots, C_{m_i}^{(i)} \rbrace $ represent the two samples for $i=1,2$, and let $\mathcal{C} = \mathcal{C}_1 \cup \mathcal{C}_2$ denote the combined dataset with a total size of $m_1 + m_2$. Denote $T(\cdot, \cdot)$ as the mechanism for computing the test statistic for one of the four tests. The pipeline proceeds as follows:
\begin{enumerate}
	\item Calculate the observed test statistic $\hat{T}_{m_1, m_2} = T(\mathcal{C}_1, \mathcal{C}_2)$ for the original data.
	\item For $n=1, \ldots, N$ iterations:
	\begin{itemize}
		\item Randomly permute the combined dataset $\mathcal{C}$.
		\item Assign $m_1$ observations to $\mathcal{C}_1^{(n)}$ and the remaining $m_2$ observations to $\mathcal{C}_2^{(n)}$.
		\item Compute the test statistic $T^{(n)} = T(\mathcal{C}_1^{(n)}, \mathcal{C}_2^{(n)})$.
	\end{itemize}
	\item Calculate the permutation $p$-value using:
	\begin{equation*}
		\hat{p} = \frac{1}{N+1} \left( \sum_{n=1}^N I(\hat{T}_{m_1, m_2} \leq T^{(n)}) + 1 \right),
	\end{equation*}
	where $I(\cdot)$ is the indicator function.
\end{enumerate}

Once a significance level $\alpha \in (0,1)$ is specified, the test based on permutation rejects the null hypothesis of equal distributions if $\hat{p} \leq \alpha$. This approach provides strong theoretical guarantees for controlling false positive rates \citep{romano_2005_ExactApproximateStepdown, you_2022_ComparingMultipleLatent}.

Permutation-based testing has some advantages. First, it is distribution-free and avoids reliance on asymptotic approximations, making it suitable for small sample sizes or data with non-standard distributions. Second, the framework is easily adaptable to different test statistics and discrepancy measures. It further accommodates complex data structures such as correlation matrices on $\CORR$.


\section{Results}\label{sec:examples}

In this section, we perform several standard learning tasks using both simulated and real datasets. The simulations are designed to validate the advantages of the newly introduced geometric structures on $\CORR$, particularly in terms of computational efficiency and the accuracy of estimating centroid measures such as the \Frechet mean and median. These benchmarks serve to highlight the practical benefits of using ECM and LEC geometries compared to traditional approaches.

We then transition to applying these methods to real-world tasks in neuroimaging, including behavior score prediction, fingerprinting, and two-sample hypothesis testing for group differences. These tasks are representative of common applications in the field, where the ability to process and analyze functional connectivity data effectively is critical. All computations were carried out on a consumer-grade laptop (MacBook Air M1, featuring an 8-core CPU and 8GB of unified memory), demonstrating the computational feasibility of the proposed methods even with modest hardware resources.

\subsection{Simulation 1. Scalable computation of \Frechet mean}

The \Frechet mean, a generalization of the standard mean from Euclidean space to manifold settings, is a critical method for summarizing a set of manifold-valued observations. In this experiment, we empirically demonstrate how the newly introduced ECM and LEC geometries enhance the scalability of \Frechet mean computation for large correlation matrices, which correspond to the scale of functional parcellation typically encountered in modern neuroimaging studies.

Given the limited theoretical understanding of probability distributions for correlation matrices, we employed the Wishart distribution $\mathcal{W}_p(V, n)$ to generate $(p \times p)$ covariance matrices. The Wishart distribution allows for the generation of positive-definite matrices, with $V \in \mathcal{S}_{++}^p$ as the scale matrix and $n$ as the degrees of freedom. To adapt these samples for analysis on $\CORR$, we applied a normalization process. This adjustment is necessary because Wishart-distributed samples are scatter matrices, essentially constant multiples of empirical covariance matrices, which do not inherently satisfy the unit diagonal constraint of $\CORR$. By normalizing the samples, we ensured adherence to the geometric constraints of $\CORR$, preserving their suitability for manifold-based computations. For the experiment, we selected an identity matrix $I_p$ as the scale matrix and set the degrees of freedom to $n = 2p$ for each dimension $p \in \lbrace 100, 200, \ldots, 1000 \rbrace$. This choice of $n$ avoids rank deficiency in the generated samples, ensuring full-rank correlation matrices suitable for downstream analysis. From each dimensional setting, 100 correlation matrices were randomly sampled and used as inputs for \Frechet mean computation.

\begin{figure}[t!]
	\centering
	\includegraphics[width=.5\linewidth]{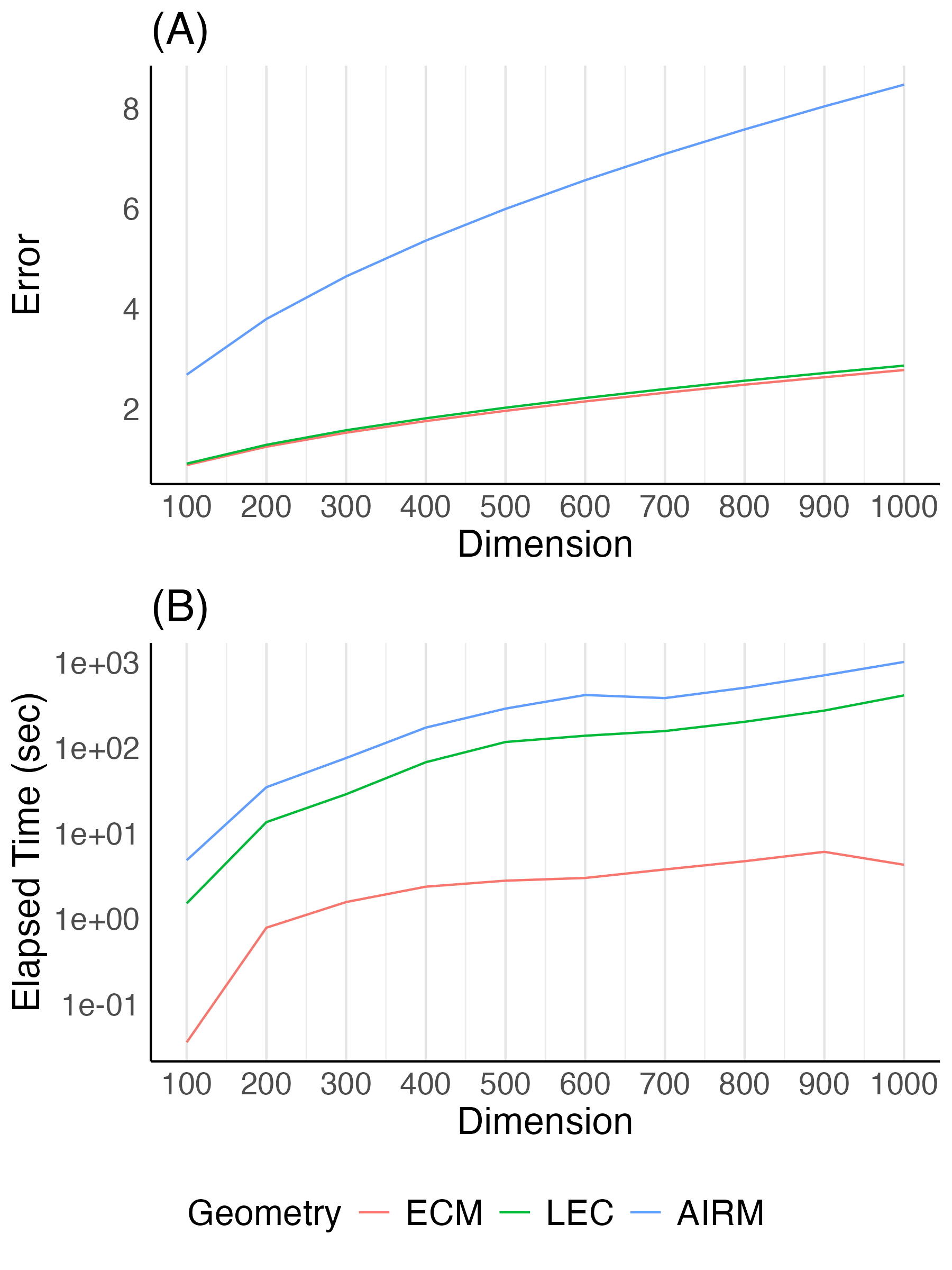}
	\caption{Simulation results for the \Frechet mean computation. For varying dimensions, the \Frechet mean was estimated for a random sample of 100 correlation matrices drawn from the Wishart distribution. Performance under different geometries is reported in terms of (A) the magnitude of error for the estimated \Frechet mean, measured using the Frobenius norm, and (B) the elapsed computation time, measured in seconds.}
	\label{fig:sim1}
\end{figure}

The simulation results are presented in Figure \ref{fig:sim1}, summarizing the deviation of the \Frechet mean from the model scale matrix and the computation time across three competing geometries: ECM and LEC for $\CORR$, and AIRM as the ambient geometry on $\SPD$. 

Across all dimensions, ECM and LEC geometries demonstrated comparable error levels, measured as the Frobenius norm deviation of the computed \Frechet mean from the model scale matrix, relative to AIRM. This observation highlights that the newly introduced geometries are effective for analyzing correlation-valued data, even in the context of straightforward centroid computations. Furthermore, ECM and LEC geometries required significantly less computational time than AIRM, with ECM outperforming LEC. This performance difference is naturally expected, as LEC involves an additional matrix logarithm computation on top of the normalized Cholesky decomposition $\Theta$.

We conclude this section with two key observations. First, the estimation error increases as the dimensionality grows. This pattern is consistent with known challenges in modern statistics that higher-dimensional spaces typically demand larger sample sizes for accurate estimation of means, reflecting the curse of dimensionality \citep{buhlmann_2011_StatisticsHighDimensionalData}. 

Second, we did not include performance metrics for the QAM geometry due to its computational infeasibility in this setting. Under similar algorithmic conditions (e.g., identical stopping criteria), a single run of \Frechet mean computation at $p=100$ took over an hour, with computation time increasing exponentially as the dimension increased. This limitation stems from the fundamental inefficiency of QAM geometry, where most basic operations require solving large-scale nonlinear optimization problems, making it impractical for high-dimensional correlation matrices. These results underline the advantages of ECM and LEC geometries in terms of both computational efficiency and scalability.

\subsection{Simulation 2. Robust estimation of the centroid via \Frechet median}

As discussed in Section \ref{sec:methods}, the \Frechet median, which minimizes Equation \eqref{eq:frechet-general} with respect to order 1, provides a robust alternative to the \Frechet mean in the presence of outliers \citep{huber_1981_RobustStatistics}. To validate its robustness as a central tendency measure, we conducted an experiment to assess its performance under controlled contamination scenarios. Similar to the previous simulation, we generated a random sample of 100 correlation matrices at a fixed dimension of $p=300$, selected as a proxy for commonly used cortical parcellations in neuroimaging studies. Two data-generating mechanisms were employed to represent signal and noise distributions, respectively.

For the signal, the correlation matrices were drawn from the Wishart distribution with an identity scale matrix, defined as 
\[
V_{\textrm{signal}}(i,j) = \mathbb{I}\lbrace i=j\rbrace,
\]
where \( \mathbb{I} \) denotes the indicator function. For the noise, samples were drawn from the Wishart distribution with a covariance matrix derived from an autoregressive process of order 1 (AR(1)). Specifically, the AR(1) scale matrix \( V_{\textrm{noise}} \in \mathbb{R}^{300 \times 300} \) is defined as:
\begin{equation*}
	V_{\textrm{noise}}(i,j) = \rho^{|i-j|}, \quad i,j = 1, \ldots, 300,\end{equation*}
with a decay parameter \( \rho = 0.9 \), chosen to ensure a clear distinction between the signal and noise distributions.

To simulate varying levels of contamination, we generated mixed samples comprising \( k \in \lbrace 5, 10, 15, 20, 25 \rbrace \) noise samples from the AR(1) distribution and \( (100-k) \) signal samples from the identity distribution. For each contamination level, the \Frechet median was computed under the ECM, LEC, and AIRM geometries.

\begin{figure}[ht!]
	\centering
	\includegraphics[width=.5\linewidth]{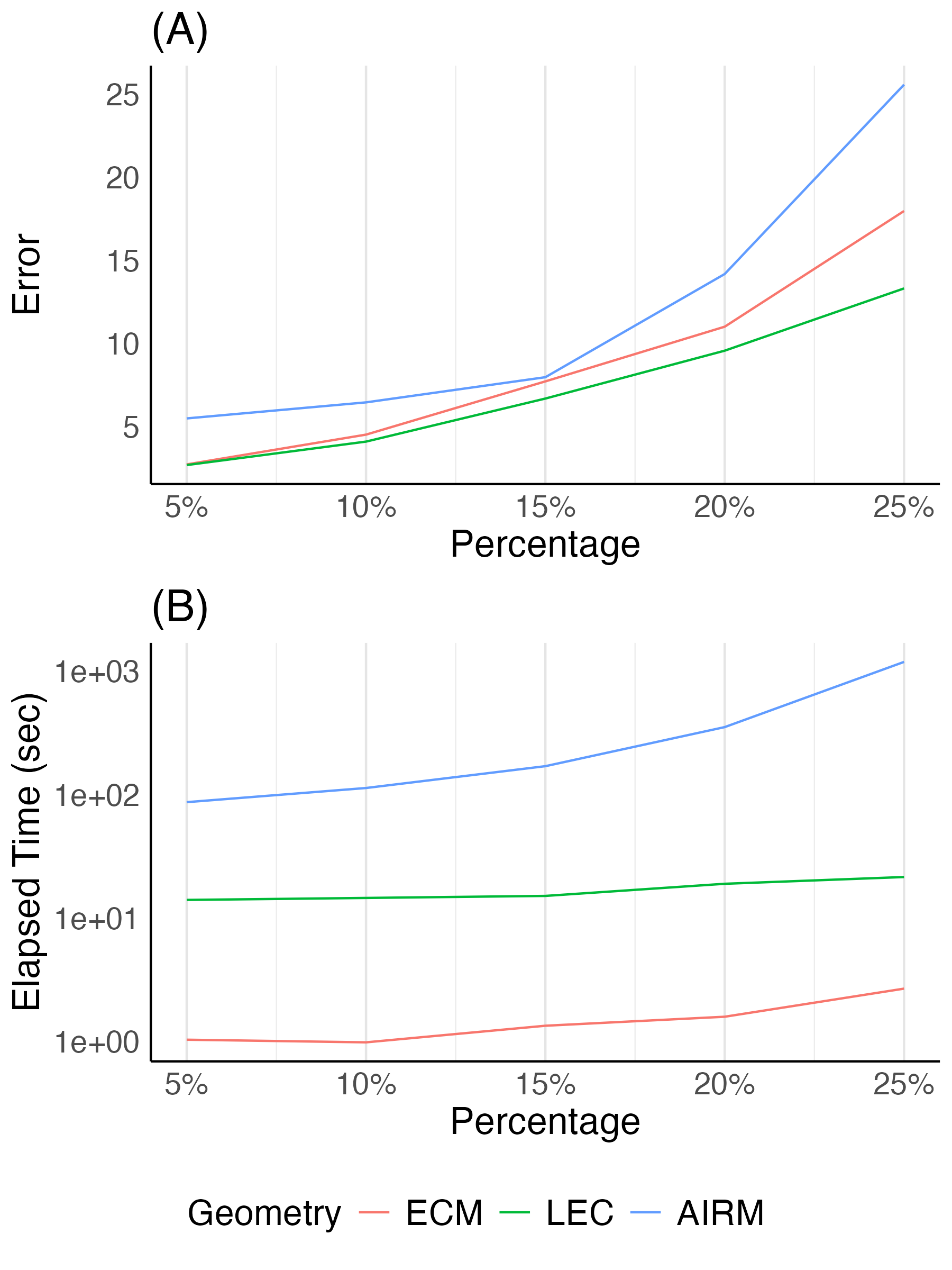}
	\caption{Simulation results for the \Frechet median example. For each contamination level, a random sample composed of two types of correlation matrices is drawn, and the \Frechet median is computed. The performance of different geometries is shown in terms of (A) the error magnitude, measured as the Frobenius norm difference between the estimated \Frechet median and the identity matrix, and (B) elapsed computation time, measured in seconds.}
	\label{fig:sim2}
\end{figure}

The results of the experiment are summarized in Figure \ref{fig:sim2}. As expected, the error increased with higher contamination levels, reflecting the challenge of accurately estimating a robust centroid as the proportion of noise samples grows. Nevertheless, the two correlation-specific geometries, ECM and LEC, consistently exhibited smaller error magnitudes compared to AIRM, demonstrating their robustness and effectiveness in handling noisy data.

In terms of computational efficiency, both ECM and LEC significantly outperformed AIRM, reducing computation times by at least an order of magnitude. The computational inefficiency of AIRM arises from the numerical complexity of operations required in SPD geometry. Conversely, ECM and LEC geometries map correlation matrices into vectorized forms, allowing for the efficient application of the Weiszfeld algorithm \citep{weiszfeld_1937_PointPourLequel} for median estimation. Notably, ECM demonstrated superior computational performance compared to LEC at all levels of noise contamination, with the advantage becoming more pronounced as the number of noise samples increased.

The relatively consistent runtime of LEC across contamination levels is attributed to the matrix logarithm step, which dominates its overall computational cost. This step minimizes the impact of additional noise samples on runtime, unlike ECM, where the presence of outliers introduces greater complexity to the iterative centroid estimation process.

\begin{figure*}[ht!]
	\centering \includegraphics[width=.95\linewidth]{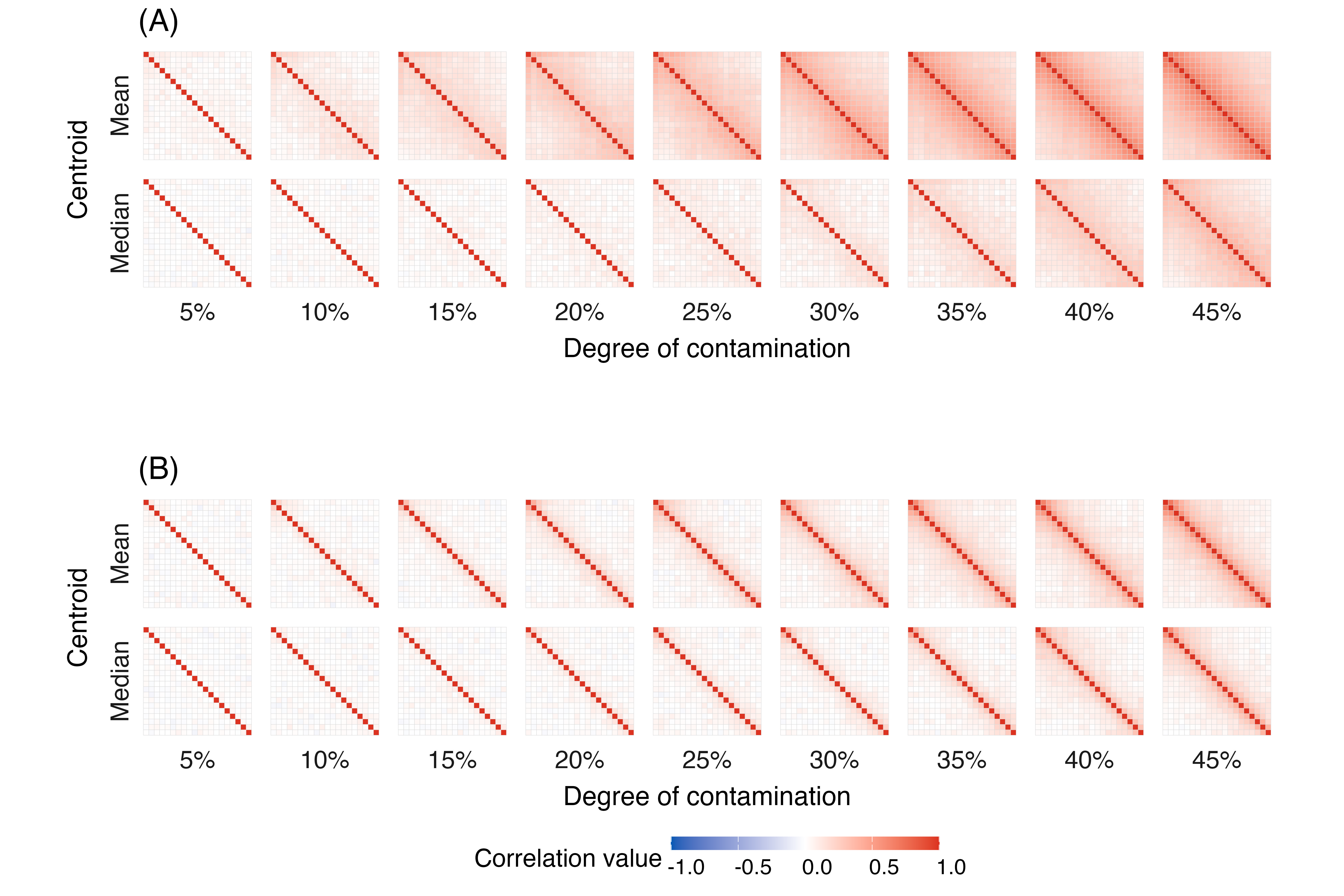} 
	\caption{ Visualization of two centroids, the \Frechet mean and the \Frechet median, in the reduced dimension under (A) the Euclidean-Cholesky metric and (B) the Log-Euclidean Cholesky metric. A random sample of 100 correlation matrices was drawn, where $k$ of them were from a Wishart distribution (noise) and the remaining $100-k$ were identity matrices. All matrices were generated in accordance with the perturbation scheme described in the text. } \label{fig:meanmedian}
\end{figure*}

Finally, we modified the experiment to demonstrate that the \Frechet median produces a robust centroid under both geometries. For visualization, we reduced the dimensionality to $p=20$ and generated a random sample of 100 correlation matrices composed of two types: perturbed versions of the identity (signal) and AR(1) scale (noise). As shown in Figure~\ref{fig:meanmedian}, the \Frechet median remains relatively robust at higher contamination levels, because the off-diagonal elements of the centroids exhibit smaller magnitudes in both the ECM and LEC geometries. This supports the observation that the median, as a robust alternative to the mean, retains its robustness on the correlation manifold.

In summary, this experiment highlights the robustness and computational efficiency of ECM and LEC geometries for \Frechet median computation in the presence of noise. Both geometries demonstrated superior accuracy compared to AIRM, with ECM providing the best overall balance of error performance and runtime efficiency. These findings underscore the practicality of ECM and LEC geometries for robust statistical learning tasks involving correlation matrices, especially in high-dimensional applications.

\subsection{Real data analysis}

For the experimental analysis, we utilized two publicly available datasets to evaluate the performance of the proposed methods across a variety of tasks. 

The first dataset was drawn from the 1200-subject release of the Human Connectome Project (HCP) database\footnote{\url{https://www.humanconnectome.org/}} \citep{vanessen_2012_HumanConnectomeProject}. From this cohort, we selected 980 subjects (Age: $28.71 \pm 3.71$ years, range 22–37; Males: 460, Females: 520). Each subject completed two 15-minute resting-state fMRI recordings with left-to-right (LR) and right-to-left (RL) phase encoding, resulting in four sessions per subject. The time-series data were sampled at 0.72 Hz, with 1200 time points per session. We used the version of the extensively processed fMRI data where preprocessing followed the HCP minimal preprocessing pipeline and mapped the data onto cortical surfaces \citep{glasser_2013_MinimalPreprocessingPipelines}. Additional cleaning was performed using the HCP ICA-FIX pipeline, which regresses out motion-related artifacts and noise components identified via independent component analysis (ICA) \citep{salimi-khorshidi_2014_AutomaticDenoisingFunctional, griffanti_2014_ICAbasedArtefactRemoval}.

For network-level functional connectivity analysis, time series data were extracted using the Schaefer atlas \citep{schaefer_2018_LocalGlobalParcellationHuman}, which parcellates the cortical surface into 300 regions of interest (ROIs). Principal component analysis (PCA) was applied to the time series within each ROI, with the first principal component used to summarize the BOLD signal in each region. Empirical correlation matrices based on this parcellation were frequently rank-deficient due to the high number of ROIs relative to the available time points. To mitigate this, we applied three covariance matrix estimators: (1) Oracle Approximating Shrinkage (OAS) \citep{chen_2010_ShrinkageAlgorithmsMMSE}, (2) Ledoit-Wolf (LW) Shrinkage \citep{ledoit_2004_WellconditionedEstimatorLargedimensional}, and (3) Ridge Estimation \citep{mejia_2018_ImprovedEstimationSubjectlevel}, incorporating a regularization term with $\tau = 1.0$, such that $\Sigma_{\textrm{Ridge}} = \Sigma_\textrm{empirical} + \tau \cdot I$. All resulting covariance matrices were normalized to adhere to the constraints of $\CORR$, ensuring unit diagonals.

The second dataset was the EEG motor movement and imagery dataset \citep{schalk_2009_EEGMotorMovement}, available through the PhysioNet database\footnote{\url{https://physionet.org}} \citep{goldberger_2000_PhysioBankPhysioToolkitPhysioNet}. This dataset comprises 64-channel EEG recordings from 109 participants, collected using the BCI2000 system \citep{schalk_2004_BCI2000GeneralPurposeBrainComputer}. After excluding six participants due to annotation errors, we retained data from 103 subjects. Participants completed motor execution and motor imagery tasks involving fists and feet movements across 14 experimental sessions. Neural activity was recorded at a sampling rate of 160 Hz. For this analysis, we selected a single participant (S001) and focused on the motor imagery tasks.

Preprocessing followed the pipeline outlined in our previous study \citep{you_2022_GeometricLearningFunctional}. First, 32 channels identified as `bad' (e.g., flat signals or poor signal-to-noise ratios) were removed. A Butterworth IIR band-pass filter with cutoff frequencies at 7 Hz and 35 Hz was applied using a two-pass zero-phase method. The filtered signals were segmented into epochs from each stimulus onset to one-second post-stimulus, resulting in 161 temporal measurements per epoch. This process yielded 45 samples, with 21 corresponding to feet movements and 24 to fists. Empirical correlation matrices computed for this dataset were full-rank, eliminating the need for regularized correlation estimators.

\subsubsection{Experiment 1. Predicting Behavior Score}


The first experiment focuses on the predictive modeling of behavior scores using correlation matrices within a regression framework. Leveraging the rich data from the Human Connectome Project (HCP), we assessed the effectiveness of nonparametric regression models for correlation-valued covariates in predicting behavioral outcomes, with a particular emphasis on the Penn Matrix Test (PMAT). The PMAT24 is a cognitive assessment tool designed to measure abstract reasoning and problem-solving abilities, serving as a brief version of Raven's Progressive Matrices \citep{bilker_2012_DevelopmentAbbreviatedNineItem}. Following \cite{you_2021_RevisitingRiemannianGeometry}, we selected two outcome variables from the dataset: the number of correct responses (\textsf{PMAT\_A\_CR}) and the total number of skipped items (\textsf{PMAT\_A\_SI}), both of which serve as indicators of fluid intelligence. After excluding three subjects due to missing scores, the final sample consisted of 977 individuals. 

\begin{figure*}[ht]
	\centering
	\includegraphics[width=\linewidth]{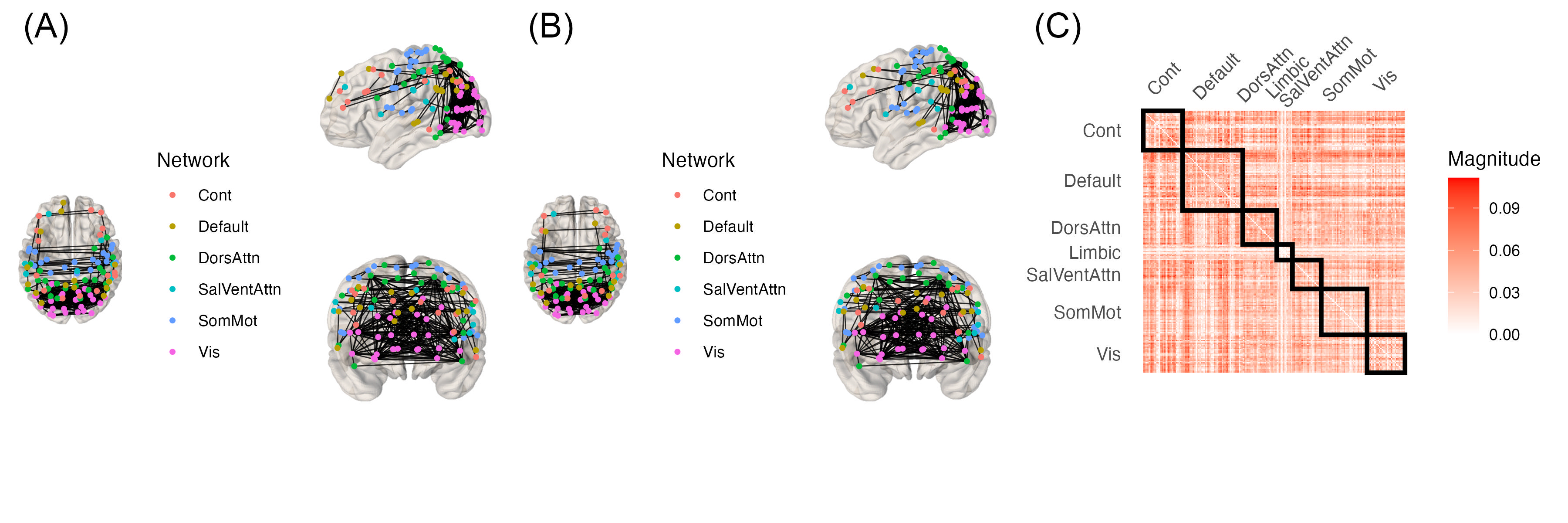}
	\caption{
		Visualization of average correlation networks estimated using oracle-approximating shrinkage estimators. The \Frechet means of subjects with (A) the top 10\% and (B) the bottom 10\% of  \textsf{PMAT\_A\_CR}  scores were computed under the Euclidean-Cholesky metric. Each correlation matrix was binarized to retain only the top 2\% of connections with the largest magnitudes. The elementwise difference matrix is shown in (C), where each entry represents the absolute difference between the two mean networks. Bounding boxes indicate the seven functional networks defined in \cite{schaefer_2018_LocalGlobalParcellationHuman}.}
	\label{fig:networks}
\end{figure*}

Before performing predictive modeling, we first took a look at potential heterogeneity within the population of correlation-based FCs. Specifically, we considered two groups of subjects stratified by their  \textsf{PMAT\_A\_CR} scores, selecting individuals from the top and bottom 10\% of the score distribution. For each subject, their FC matrix was estimated using the oracle-approximating shrinkage estimator, followed by normalization. In Figure \ref{fig:networks}, we present the \Frechet means of the correlation networks for the two groups. To highlight prominent connections, each network was binarized by retaining only the top 2\% of edges with the largest absolute magnitudes. Additionally, we computed the elementwise difference matrix between the two group-level mean FCs. Rows, and equivalently columns, of this difference matrix were then sorted and grouped based on a granular parcellation of the cerebral cortex into seven functional networks, as defined by \cite{schaefer_2018_LocalGlobalParcellationHuman}. Notably, this visualization reveals distinct patterns of connectivity differences at a macroscopic level, suggesting that functional architecture varies systematically with fluid intelligence.

We compared the performance of our proposed models against two widely used approaches in neuroimaging-based predictive modeling: connectome-based predictive modeling (CPM) \citep{shen_2017_UsingConnectomebasedPredictive} and least absolute shrinkage and selection operator (LASSO) regression \citep{tibshirani_1996_RegressionShrinkageSelection}. Both methods aim to predict behavioral or cognitive outcomes using functional connectivity matrices derived from preprocessed fMRI data. For both CPM and LASSO, we optimized hyperparameters using five-fold cross-validation on the training set, which was constructed by randomly splitting the data into an 80\% training subset and a 20\% testing subset.

The CPM approach involves identifying significant brain connections by correlating them with the outcome variable, and grouping these connections into positive and negative networks. The summed strengths of these networks are used as predictors in a linear regression model to estimate the behavioral outcome. This model emphasizes interpretability, as the identified networks provide insights into functional connectivity patterns associated with the behavior in question.

LASSO regression, on the other hand, operates on a design matrix where each row represents the vectorized upper triangular part of a correlation connectome. By applying $L_1$-regularization, LASSO shrinks the coefficients of less relevant variables toward zero, effectively performing variable selection. The non-zero coefficients correspond to the most influential features, which are then used to predict the behavioral outcome. Cross-validation ensures robust hyperparameter tuning and generalizability, with feature selection repeated independently for each fold.

Both CPM and LASSO offer advantages in linking functional connectivity patterns to behavior, particularly in terms of simplicity and interpretability. CPM highlights key networks by grouping edges into interpretable positive and negative categories, while LASSO identifies specific connections with predictive significance. 

\begin{table*}[t!]
	\caption{Accuracy of fluid intelligence prediction. For each setting, the mean squared error (MSE) between the predicted and actual scores on the test data is reported. Across two correlation geometries - the Euclidean-Cholesky metric (ECM) and the Log-Euclidean Cholesky metric (LEC) - three regression models (Gaussian process regression [GP], kernel regression [KERN], and support vector regression [SVR]) were applied to correlation-valued functional connectomes estimated using the Ledoit-Wolf estimator (LW), the oracle approximating shrinkage estimator (OAS), and the $L_2$-regularized estimator (Ridge) with a penalty $\tau=1$.}
	
	\label{table:regression}
	\centering
	\begin{tabular}{|c|c|c|c|c|c|c|c|c|c|}
		\hline
		\multirow{3}{*}{Outcome} & \multirow{3}{*}{\begin{tabular}{@{}c@{}}Correlation \\ estimator\end{tabular}}  & \multicolumn{6}{c|}{Geometry}                                                                                                              & \multirow{3}{*}{\begin{tabular}{@{}c@{}}LASSO \\ Regression\end{tabular}} & \multirow{3}{*}
		{CPM}\\ \cline{3-8}
		&                           & \multicolumn{3}{c|}{ECM}                                                       & \multicolumn{3}{c|}{LEC}                                  &                         &                      \\ \cline{3-8}
		&                           & \multicolumn{1}{c|}{GP} & \multicolumn{1}{c|}{KERN} & \multicolumn{1}{c|}{SVR} & \multicolumn{1}{c|}{GP} & \multicolumn{1}{c|}{KERN} & SVR &                         &                      \\ \hline \hline 
		\multirow{3}{*}{ \textsf{PMAT\_A\_CR}} 
		& LW & 4.50 & 4.60 & 4.41 & 4.50 & 4.60 & 4.44 & 13.29 & 32.33  \\ \cline{2-10} 
		&OAS & 4.50 & 4.56 & 4.50 & 4.50 & 4.43 & 4.45 & 12.58 & 28.00 \\ \cline{2-10} 
		&Ridge & 4.50 & 4.56 & 4.50 & 4.50 & 4.59 & 4.50 & 9.41 & 28.73 \\ \hline
		\multirow{3}{*}{ \textsf{PMAT\_A\_SI}}
		& LW & 3.69 & 3.90 & 3.66 & 3.69 & 3.79 & 3.69 & 5.46 & 28.36 \\  \cline{2-10} 
		& OAS & 3.69 & 3.91 & 3.69 & 3.69 & 4.13 & 3.69 & 11.58 & 27.06 \\   \cline{2-10} 
		& Ridge & 3.69 & 3.75 & 3.69 & 3.69 & 3.82 & 3.69 & 8.57 & 26.53 \\  \hline
	\end{tabular}
\end{table*}

Table \ref{table:regression} summarizes the accuracy results for the predictive modeling task. It is immediately apparent that the nonparametric regression models leveraging the newly introduced geometries outperform the two competing models across all outcomes and estimators. This finding aligns with expectations, as nonlinear methods generally offer greater flexibility in capturing complex relationships between covariates and outcomes, albeit with a trade-off in interpretability. The results, derived from test data errors, provide empirical evidence supporting the superior predictive power of the proposed regression framework utilizing the two alternative geometries of $\CORR$.

A noteworthy observation is that CPM consistently demonstrated the poorest performance, even when compared to the basic LASSO model applied to half-vectorized covariates. For CPM, we employed default options for training and cross-validation and observed that the results varied significantly based on hyperparameter configurations. Methodologically, CPM identifies covariates highly correlated with the outcome and aggregates them into a single predictive variable. This approach assigns uniform weights to the selected variables while setting the coefficients of all others to zero, akin to a standard linear regression model. However, the effectiveness of this strategy becomes questionable when applied to the large number of highly correlated covariates ($\mathcal{O}(n^2)$) inherent in correlation matrices. This limitation likely stems from the well-documented challenges of constructing linear regression models with highly correlated variables, highlighting potential drawbacks of CPM in this context.

\subsubsection{Experiment 2. Fingerprinting}

Next, we evaluate the effectiveness of the novel geometric structures in the task of functional connectome fingerprinting \citep{finn_2015_FunctionalConnectomeFingerprinting}, which aims to capture individual variability in functional connectivity (FC) profiles. The fingerprinting task is formulated as follows: for 
$M$ subjects, each undergoes two independent brain scan sessions, Session 1 and Session 2, producing corresponding FC representations. Given an individual's FC from Session 2, without identifying information, the goal is to determine which subject it belongs to by comparing it to the FCs from Session 1 based on a measure of similarity. This task can be viewed as a multiclass classification problem where each class contains a single observation. A 1-nearest neighbor (1-NN) classification method is naturally employed to assign the label of the most similar subject to the test sample. Identification accuracy is calculated as:
\begin{equation*} 
	I_{\text{acc}} = \frac{\text{the number of correctly identified subjects}}{\text{the total number of subjects}}, 
\end{equation*}
where $I_{\text{acc}}$ ranges from 0 to 1, with higher values indicating better identification accuracy.

In our experiment, we selected 100 unrelated subjects from the Human Connectome Project (HCP) dataset. Time series data were extracted from all four sessions of resting-state scans, and correlation-based FC representations were constructed for each session using three different estimators: the oracle approximating shrinkage (OAS) estimator, the Ledoit-Wolf (LW) estimator, and the $L_2$-regularized Ridge estimator. This procedure resulted in four distinct FCs for each subject per estimator. The experimental design considered 12 combinations of session pairs, where one session served as the training data and another as the test data. To streamline the reporting, symmetric pairs, such as (Session 1, Session 2) and (Session 2, Session 1), were treated as equivalent. Their identification accuracy scores were averaged, reducing the number of reportable cases to six.

For comparison, we employed the similarity-based fingerprinting approach proposed by \cite{finn_2015_FunctionalConnectomeFingerprinting}. This method identifies the subject corresponding to a query by finding the individual with the maximum Pearson correlation coefficient between vectorized FCs. From a machine learning perspective, this approach aligns with a 1-NN classification model where vectorized FCs serve as the data and Pearson correlation defines the similarity metric. In our adaptation, we retained the identical experimental pipeline but replaced the definition of data and similarity with the correlation-based FCs and the geometric structures introduced previously. This modification allowed us to directly assess how the proposed geometries impact the accuracy of functional connectome fingerprinting, providing insights into their utility for capturing individual-specific patterns in FC.

\begin{figure*}[t!]
	\centering
	\includegraphics[width=.99\linewidth]{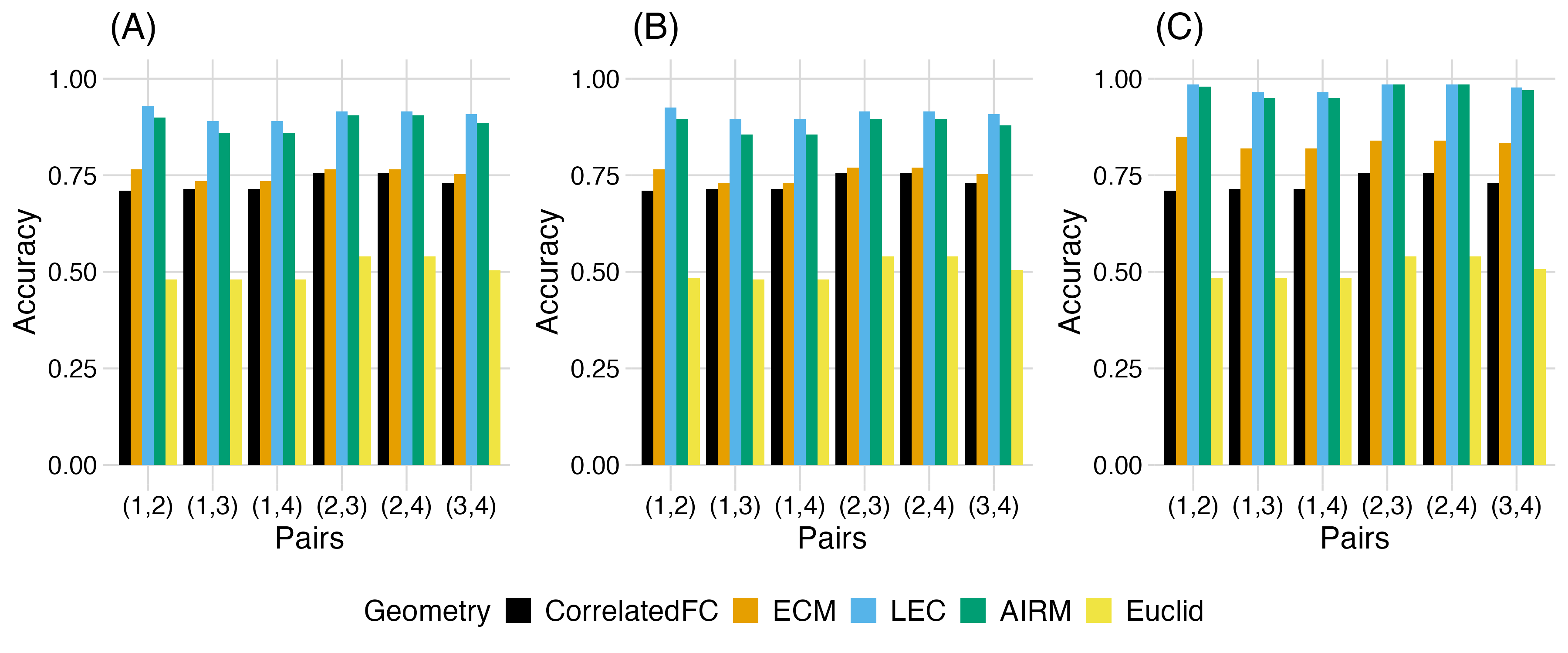}
	\caption{
		Results for the fingerprinting example with different estimators: (A) Ledoit-Wolf (LW), (B) Oracle Approximating Shrinkage (OAS), and (C) Ridge. Identification rates for six pairs of runs are reported, where each `pair' represents the average of two runs with flipped training and test dataset indices.}
	\label{fig:fingerprinting}
\end{figure*}

Figure \ref{fig:fingerprinting} summarizes the experimental results. The patterns observed across different pairs of runs are consistent, indicating low heterogeneity in the relative association of FCs across sessions for identification purposes. Across all estimators, the baseline method outperformed naive identification based on the Euclidean distance between FCs, which aligns with prior expectations and the findings of \cite{finn_2015_FunctionalConnectomeFingerprinting}. Notably, the incorporation of appropriate geometric structures into the space of FCs significantly enhanced performance. While the ECM geometry provided marginal improvements (except in the case of the Ridge estimator), the LEC geometry demonstrated substantial gains over the baseline method, with identification rates rising from below 75\% to approximately 90\% for both the OAS and LW estimators and even higher for the Ridge estimator.

These findings offer strong empirical support for the effectiveness of novel geometries on the correlation manifold in achieving fine-grained classification tasks. Additionally, it is worth noting the near-optimal performance of AIRM, which was comparable to LEC. AIRM, as a geometric structure on $\SPD$, serves as a valid distance metric for correlation matrices and has been shown to perform well in similar tasks \citep{abbas_2021_GeodesicDistanceOptimally}. In this study, AIRM's performance closely approached that of LEC, differing only slightly. However, this does not diminish the importance of our proposed framework. For example, AIRM does not preserve the correlation structure during operations such as mean computation, which limits its utility in specific scenarios compared to the proposed geometries.

\subsubsection{Experiment 3. Hypothesis Testing}

The final experiment focuses on two-sample hypothesis testing to determine whether two sets of correlations share the same underlying distribution, using an EEG dataset as the basis for analysis. For each 32-channel signal, after removing any bad channels, three types of functional connectivity representations were computed: the LW estimator, the OAS estimator, and the sample correlation matrix (SCM). After normalization, these representations served as inputs for the hypothesis testing framework, enabling a robust assessment of distributional equivalence between the two sets of correlations.

\begin{figure*}[ht!]
	\centering
	\includegraphics[width=.99\linewidth]{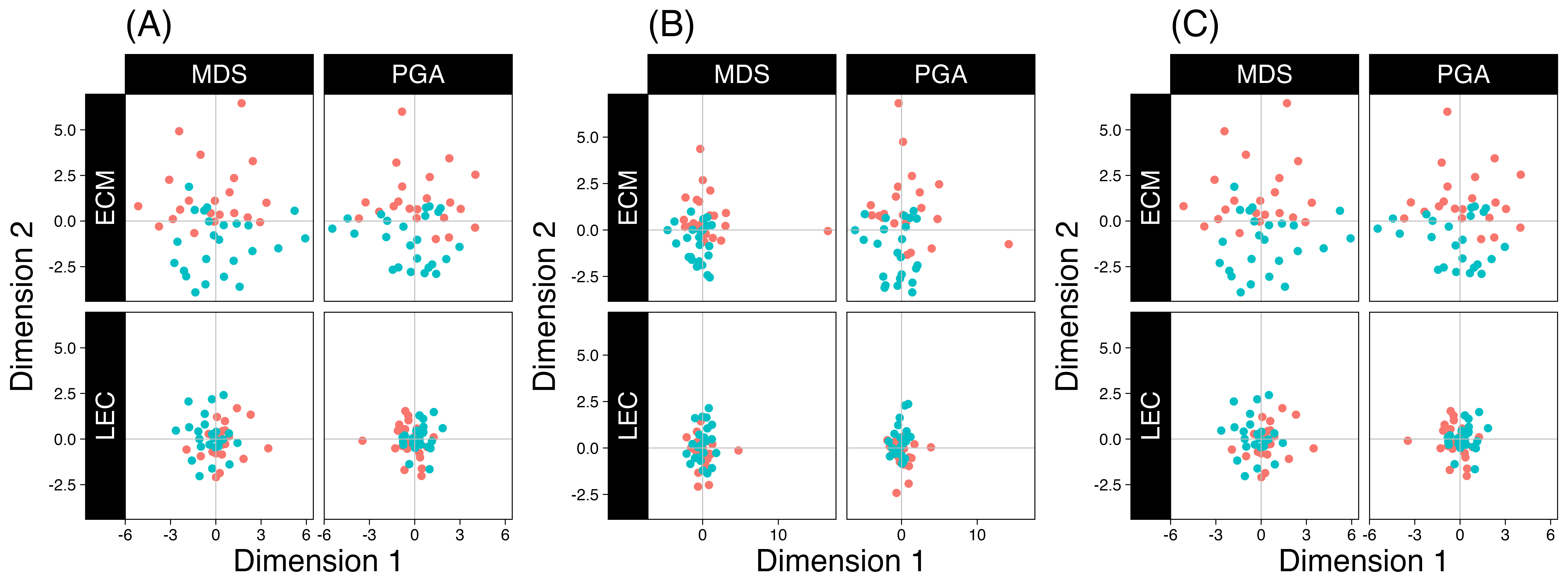}
	\caption{Low-dimensional embedding of EEG hypothesis testing data with different estimators in $\mathbb{R}^2$: (A) Ledoit-Wolf (LW), (B) Oracle Approximating Shrinkage (OAS), and (C) empirical correlation matrix (SCM). Each subplot presents embeddings generated by a combination of different geometries and algorithms.}
	\label{fig:eeg_visualization}
\end{figure*}

Before conducting the hypothesis testing, we visualized the data distributions by projecting them into a two-dimensional space using various geometries and algorithms, as illustrated in Figure \ref{fig:eeg_visualization}. The resulting visualizations reveal noticeable differences in the data distribution shapes depending on the chosen geometry. With the ECM geometry, all estimators exhibited some degree of separation between the two classes, suggesting distinguishable patterns in the data. In contrast, visualizations based on the LEC geometry showed nearly overlapping distributions for the two sets of correlations, indicating reduced separability. This stark contrast between the geometries underscores their significant influence on data representation and the subsequent interpretability of results.

\begin{table*}[h]
	\caption{Empirical $p$-values for the hypothesis testing example with EEG data. (* $p < 0.05$, ** $p < 0.01$)}
	\label{table:testing}
	\centering
	\begin{tabular}{|cc|ccc|ccc|}
		\hline
		\multicolumn{2}{|c|}{Geometry}                      & \multicolumn{3}{c|}{ECM}                                           & \multicolumn{3}{c|}{LEC}                                           \\ \hline
		\multicolumn{2}{|c|}{Estimator}                     & \multicolumn{1}{c|}{LW}     & \multicolumn{1}{c|}{OAS}    & SCM    & \multicolumn{1}{c|}{LW}     & \multicolumn{1}{c|}{OAS}    & SCM    \\ \hline\hline
		\multicolumn{1}{|c|}{\multirow{3}{*}{Tests}} & Biswas-Ghosh   & \multicolumn{1}{c|}{0.0149*} & \multicolumn{1}{c|}{0.0197*} & 0.0133* & \multicolumn{1}{c|}{0.3729} & \multicolumn{1}{c|}{0.2831} & 0.5911 \\ \cline{2-8} 
		\multicolumn{1}{|c|}{}                       & MMD  & \multicolumn{1}{c|}{0.4994} & \multicolumn{1}{c|}{0.2358} & 1      & \multicolumn{1}{c|}{0.1}    & \multicolumn{1}{c|}{0.21}   & 0.6804 \\ \cline{2-8} 
		\multicolumn{1}{|c|}{}                       & Wasserstein & \multicolumn{1}{c|}{0.0002**} & \multicolumn{1}{c|}{0.0002**} & 0.0004** & \multicolumn{1}{c|}{0.0027**} & \multicolumn{1}{c|}{0.0007**} & 0.0118* \\ \hline
	\end{tabular}
\end{table*}

Next, we performed hypothesis testing to evaluate the equality of distributions using three proposed tests under both geometries for all considered estimators. The number of resampling iterations was set to \(10^4 - 1\), which is adequate for the size of our dataset. The empirical \(p\)-values obtained are summarized in Table \ref{table:testing}. Notably, the Biswas-Ghosh test rejected the null hypothesis of equal distributions for all estimators under the ECM geometry, whereas the LEC geometry showed no statistically significant differences. This result aligns with the prior visualizations, reinforcing the conclusion that the choice of geometry strongly influences both low-dimensional embeddings and statistical outcomes.

For the other tests, based on MMD and Wasserstein distance, the results were mixed, with some cases rejecting the null hypothesis and others failing to do so. While this variability might appear unsatisfactory, it highlights critical considerations for practitioners. The MMD test, which relies on kernel methods, is sensitive to the choice of kernel and its parameters. We used the squared exponential kernel, as discussed in Proposition \ref{thm:pdkernel}, with a default parameter value of \(\theta=1\). This choice likely led to conservative results, as optimal performance requires careful tuning of \(\theta\), which controls the penalization of distant observations. Adjusting \(\theta\) to better suit the data could improve the sensitivity of the test.

In contrast, the Wasserstein distance-based test rejected the null hypothesis for all combinations. However, the \(p\)-values obtained from the LEC geometry were notably higher than those from the ECM geometry, indicating that the test detected smaller discrepancies under the LEC geometry. The consistent rejection of the null hypothesis across all combinations may be attributed to challenges in estimating the Wasserstein distance in high-dimensional settings \citep{panaretos_2020_InvitationStatisticsWasserstein}. The EEG dataset used in this experiment consists of signals from 32 channels, resulting in an intrinsic dimensionality of 496 for the correlation FCs, while the sample size is only 45. This imbalance between dimensionality and sample size likely complicates the estimation process and hinders the ability to draw robust statistical inferences, as evidenced by the observed outcomes.

\section{Discussion}\label{sec:discussion}

The correlation matrix, a fundamental tool in functional network analysis, encapsulates collective information that extends beyond independent pairwise correlation coefficients. As such, treating it as a manifold-valued object with distinct geometric structures is a natural and advantageous perspective. 

Recent work, such as \cite{you_2022_GeometricLearningFunctional}, has integrated the quotient geometry of the correlation manifold into machine learning and statistical inference, enabling more robust functional connectivity analysis. However, challenges like computational inefficiency and instability in high-dimensional settings with many ROIs have limited its practical use.

To overcome these limitations, we introduced alternative geometric characterizations of the correlation manifold, as proposed by \cite{thanwerdas_2022_TheoreticallyComputationallyConvenient}. These alternatives offer dual advantages: they enable the application of well-established learning algorithms in traditional Euclidean settings and provide substantial computational benefits over the quotient geometry. Leveraging these advancements, we implemented a suite of computational operations on the correlation manifold, including measures of central tendency, cluster analysis, hypothesis testing, and low-dimensional embedding. These tools are particularly advantageous for large-scale functional network analyses, where the brain is typically divided into hundreds of regions. Consequently, we proposed new techniques for functional connectivity and statistical learning aimed at population-level inference. The efficacy of these algorithms, grounded in the novel geometric structures, was validated using both simulated and real datasets, encompassing a variety of common neuroimaging analysis tasks.

Despite these contributions, some questions remain unresolved, opening avenues for future exploration. One critical issue is the selection of an appropriate geometry. While our study highlighted the comparative advantages of ECM and LEC geometries over QAM and other ambient geometries, no single geometry emerged as universally superior. In practical applications, it is often impossible to determine the optimal model in advance, and theoretical guarantees are lacking. This makes geometry selection a hyperparameter tuning problem, best addressed through data-driven approaches such as cross-validation.

Beyond the coverage of this paper, these contributions can be applied to a broader range of problems. For instance, multi-site harmonization of functional connectivity, recently approached from a Riemannian geometric perspective, may benefit from the computational efficiencies introduced by our framework. Techniques such as replacing centroids with the \Frechet mean and translation operations with parallel transport have been demonstrated on the SPD manifold \citep{simeon_2022_RiemannianGeometryFunctional, honnorat_2024_RiemannianFrameworksHarmonization}. Our work could facilitate the efficient application of such methods to correlation-valued functional connectivity. Additionally, modeling continuous trajectories of dynamic functional connectivity \citep{preti_2017_DynamicFunctionalConnectome}, a task that typically relies on window-based segmentation, could be enhanced through manifold-valued regression. As demonstrated in \cite{you_2021_RevisitingRiemannianGeometry}, nonparametric regression on correlation manifolds could provide a powerful approach, particularly when computational efficiency is ensured through appropriate geometric structures.

The potential applications of these computational advancements extend well beyond neuroscience. In finance, for example, correlation matrices have long been used to model associations among asset returns, supporting tasks such as portfolio optimization, risk assessment, and change-point detection \citep{mantegna_1999_HierarchicalStructureFinancial, bonanno_2003_TopologyCorrelationbasedMinimal, onnela_2003_DynamicsMarketCorrelations}. Similarly, climate science could benefit from geometric analysis of correlation matrices to study interdependencies among climate variables over spatial and temporal scales \citep{braunisch_2013_SelectingCorrelatedClimate, runge_2014_QuantifyingStrengthDelay}. The development of scalable and efficient computational pipelines for correlation matrix analysis thus holds promise for diverse fields.

To encourage broader adoption and foster further exploration, we have consolidated all the algorithms discussed in this paper into a MATLAB toolbox, \textsf{CORRbox}, available on GitHub (\url{https://github.com/kisungyou/CORRbox}). By providing an accessible and optimized platform, we aim to democratize the analysis of functional networks and inspire the integration of geometric approaches to correlation matrix analysis in a wide array of scientific disciplines. Future research could expand this foundation by exploring multi-modal integration, real-time analysis pipelines, and deeper theoretical characterizations of manifold structures to further enhance the capabilities and applications of these tools.

\section*{Acknowledgement}
Kisung You was supported by a PSC-CUNY Award (TRADB-55-511), jointly funded by The Professional Staff Congress and The City University of New York. Hae-Jeong Park was supported by the Bio\&Medical Technology Development Program of the National Research Foundation (NRF) funded by the Korean government (MSIT) (No. RS-2024-00401794).

\bibliographystyle{dcu}
\bibliography{references}

@article{eaves_1994_SubtypesAutismCluster,
	title = {Subtypes of autism by cluster analysis},
	volume = {24},
	copyright = {http://www.springer.com/tdm},
	issn = {0162-3257, 1573-3432},
	url = {http://link.springer.com/10.1007/BF02172209},
	doi = {10.1007/BF02172209},
	language = {en},
	number = {1},
	urldate = {2024-12-28},
	journal = {Journal of Autism and Developmental Disorders},
	author = {Eaves, Linda C. and Ho, Helena H. and Eaves, David M.},
	month = feb,
	year = {1994},
	pages = {3--22},
}

@article{baldi_1989_NeuralNetworksPrincipal,
	title = {Neural networks and principal component analysis: {Learning} from examples without local minima},
	volume = {2},
	copyright = {https://www.elsevier.com/tdm/userlicense/1.0/},
	issn = {08936080},
	shorttitle = {Neural networks and principal component analysis},
	url = {https://linkinghub.elsevier.com/retrieve/pii/0893608089900142},
	doi = {10.1016/0893-6080(89)90014-2},
	language = {en},
	number = {1},
	urldate = {2024-12-28},
	journal = {Neural Networks},
	author = {Baldi, Pierre and Hornik, Kurt},
	month = jan,
	year = {1989},
	pages = {53--58},
}

@article{runge_2014_QuantifyingStrengthDelay,
	title = {Quantifying the {Strength} and {Delay} of {Climatic} {Interactions}: {The} {Ambiguities} of {Cross} {Correlation} and a {Novel} {Measure} {Based} on {Graphical} {Models}},
	volume = {27},
	issn = {0894-8755, 1520-0442},
	shorttitle = {Quantifying the {Strength} and {Delay} of {Climatic} {Interactions}},
	url = {http://journals.ametsoc.org/doi/10.1175/JCLI-D-13-00159.1},
	doi = {10.1175/JCLI-D-13-00159.1},
	language = {en},
	number = {2},
	urldate = {2024-12-27},
	journal = {Journal of Climate},
	author = {Runge, Jakob and Petoukhov, Vladimir and Kurths, Jürgen},
	month = jan,
	year = {2014},
	pages = {720--739},
}

@article{braunisch_2013_SelectingCorrelatedClimate,
	title = {Selecting from correlated climate variables: a major source of uncertainty for predicting species distributions under climate change},
	volume = {36},
	issn = {0906-7590, 1600-0587},
	shorttitle = {Selecting from correlated climate variables},
	url = {https://onlinelibrary.wiley.com/doi/10.1111/j.1600-0587.2013.00138.x},
	doi = {10.1111/j.1600-0587.2013.00138.x},
	language = {en},
	number = {9},
	urldate = {2024-12-27},
	journal = {Ecography},
	author = {Braunisch, Veronika and Coppes, Joy and Arlettaz, Raphaël and Suchant, Rudi and Schmid, Hans and Bollmann, Kurt},
	month = sep,
	year = {2013},
	pages = {971--983},
}

@book{mohri_2012_FoundationsMachineLearning,
	address = {Cambridge, MA},
	series = {Adaptive computation and machine learning series},
	title = {Foundations of machine learning},
	isbn = {978-0-262-01825-8},
	publisher = {MIT Press},
	author = {Mohri, Mehryar and Rostamizadeh, Afshin and Talwalkar, Ameet},
	year = {2012},
}

@book{berg_1984_HarmonicAnalysisSemigroups,
	address = {New York, NY},
	series = {Graduate {Texts} in {Mathematics}},
	title = {Harmonic {Analysis} on {Semigroups}},
	volume = {100},
	copyright = {http://www.springer.com/tdm},
	isbn = {978-1-4612-7017-1 978-1-4612-1128-0},
	urldate = {2024-09-12},
	publisher = {Springer New York},
	author = {Berg, Christian and Christensen, Jens Peter Reus and Ressel, Paul},
	year = {1984},
}

@book{bhatia_2009_PositiveDefiniteMatrices,
	title = {Positive {Definite} {Matrices}},
	isbn = {978-1-4008-2778-7},
	shorttitle = {Positive {Definite} {Matrices}},
	url = {https://www.degruyter.com/document/doi/10.1515/9781400827787/html},
	urldate = {2021-09-27},
	publisher = {Princeton University Press},
	author = {Bhatia, Rajendra},
	month = dec,
	year = {2009},
	doi = {10.1515/9781400827787},
}

@article{afsari_2011_Riemannian$L_p$Center,
	title = {Riemannian \$\{{L}\}\_p\$ center of mass: {Existence}, uniqueness, and convexity},
	volume = {139},
	issn = {0002-9939},
	shorttitle = {Riemannian \${L}{\textasciicircum}\{p\}\$ center of mass},
	url = {http://www.ams.org/jourcgi/jour-getitem?pii=S0002-9939-2010-10541-5},
	doi = {10.1090/S0002-9939-2010-10541-5},
	language = {en},
	number = {02},
	urldate = {2021-09-27},
	journal = {Proceedings of the American Mathematical Society},
	author = {Afsari, Bijan},
	month = feb,
	year = {2011},
	pages = {655--655},
}

@book{absil_2008_OptimizationAlgorithmsMatrix,
	address = {Princeton, N.J. ; Woodstock},
	title = {Optimization algorithms on matrix manifolds},
	isbn = {978-0-691-13298-3},
	publisher = {Princeton University Press},
	author = {Absil, P.-A. and Mahony, R. and Sepulchre, R.},
	year = {2008},
	note = {OCLC: ocn174129993},
}

@article{abbas_2021_GeodesicDistanceOptimally,
	title = {Geodesic {Distance} on {Optimally} {Regularized} {Functional} {Connectomes} {Uncovers} {Individual} {Fingerprints}},
	volume = {11},
	issn = {2158-0014, 2158-0022},
	url = {https://www.liebertpub.com/doi/10.1089/brain.2020.0881},
	doi = {10.1089/brain.2020.0881},
	language = {en},
	number = {5},
	urldate = {2021-10-04},
	journal = {Brain Connectivity},
	author = {Abbas, Kausar and Liu, Mintao and Venkatesh, Manasij and Amico, Enrico and Kaplan, Alan David and Ventresca, Mario and Pessoa, Luiz and Harezlak, Jaroslaw and Goñi, Joaquín},
	month = jun,
	year = {2021},
	pages = {333--348},
}

@article{you_2022_RdimtoolsPackageDimension,
	title = {Rdimtools: {An} {R} package for dimension reduction and intrinsic dimension estimation},
	volume = {14},
	copyright = {All rights reserved},
	issn = {26659638},
	shorttitle = {Rdimtools},
	url = {https://linkinghub.elsevier.com/retrieve/pii/S2665963822001002},
	doi = {10.1016/j.simpa.2022.100414},
	language = {en},
	urldate = {2022-09-10},
	journal = {Software Impacts},
	author = {You, Kisung and Shung, Dennis},
	month = nov,
	year = {2022},
	pages = {100414},
}

@article{you_2022_ComparingMultipleLatent,
	title = {Comparing multiple latent space embeddings using topological analysis},
	copyright = {Creative Commons Attribution Non Commercial No Derivatives 4.0 International},
	url = {https://arxiv.org/abs/2208.12435},
	doi = {10.48550/ARXIV.2208.12435},
	urldate = {2022-09-15},
	author = {You, Kisung and Kim, Ilmun and Jin, Ick Hoon and Jeon, Minjeong and Shung, Dennis},
	year = {2022},
	note = {Publisher: arXiv
Version Number: 1},
}

@article{you_2022_GeometricLearningFunctional,
	title = {Geometric learning of functional brain network on the correlation manifold},
	volume = {12},
	copyright = {All rights reserved},
	issn = {2045-2322},
	url = {https://www.nature.com/articles/s41598-022-21376-0},
	doi = {10.1038/s41598-022-21376-0},
	language = {en},
	number = {1},
	urldate = {2022-10-22},
	journal = {Scientific Reports},
	author = {You, Kisung and Park, Hae-Jeong},
	month = oct,
	year = {2022},
	pages = {17752},
}

@article{you_2021_RevisitingRiemannianGeometry,
	title = {Re-visiting {Riemannian} geometry of symmetric positive definite matrices for the analysis of functional connectivity},
	volume = {225},
	copyright = {All rights reserved},
	issn = {10538119},
	url = {https://doi.org/10.1016/j.neuroimage.2020.117464},
	doi = {10.1016/j.neuroimage.2020.117464},
	language = {en},
	urldate = {2021-06-09},
	journal = {NeuroImage},
	author = {You, Kisung and Park, Hae-Jeong},
	month = jan,
	year = {2021},
	pages = {117464},
}

@inproceedings{yamin_2019_ComparisonBrainConnectomes,
	address = {Venice, Italy},
	title = {Comparison {Of} {Brain} {Connectomes} {Using} {Geodesic} {Distance} {On} {Manifold}: {A} {Twins} {Study}},
	isbn = {978-1-5386-3641-1},
	shorttitle = {Comparison {Of} {Brain} {Connectomes} {Using} {Geodesic} {Distance} {On} {Manifold}},
	url = {https://ieeexplore.ieee.org/document/8759407/},
	doi = {10.1109/ISBI.2019.8759407},
	urldate = {2021-10-04},
	booktitle = {2019 {IEEE} 16th {International} {Symposium} on {Biomedical} {Imaging} ({ISBI} 2019)},
	publisher = {IEEE},
	author = {Yamin, A. and Dayan, M. and Squarcina, L. and Brambilla, P. and Murino, V. and Diwadkar, V. and Sona, D.},
	month = apr,
	year = {2019},
	pages = {1797--1800},
}

@article{weiszfeld_1937_PointPourLequel,
	title = {Sur le point pour lequel la {Somme} des distances de n points donnes est minimum},
	volume = {43},
	journal = {Tohoku Mathematical Journal, First Series},
	author = {Weiszfeld, Endre},
	year = {1937},
	pages = {355--386},
}

@article{vonluxburg_2007_TutorialSpectralClustering,
	title = {A tutorial on spectral clustering},
	volume = {17},
	issn = {0960-3174, 1573-1375},
	url = {http://link.springer.com/10.1007/s11222-007-9033-z},
	doi = {10.1007/s11222-007-9033-z},
	language = {en},
	number = {4},
	urldate = {2021-10-02},
	journal = {Statistics and Computing},
	author = {von Luxburg, Ulrike},
	month = dec,
	year = {2007},
	pages = {395--416},
}

@incollection{varoquaux_2010_DetectionBrainFunctionalConnectivity,
	address = {Berlin, Heidelberg},
	title = {Detection of {Brain} {Functional}-{Connectivity} {Difference} in {Post}-stroke {Patients} {Using} {Group}-{Level} {Covariance} {Modeling}},
	volume = {6361},
	isbn = {978-3-642-15704-2 978-3-642-15705-9},
	language = {en},
	urldate = {2021-09-27},
	booktitle = {Medical {Image} {Computing} and {Computer}-{Assisted} {Intervention} – {MICCAI} 2010},
	publisher = {Springer Berlin Heidelberg},
	author = {Varoquaux, Gaël and Baronnet, Flore and Kleinschmidt, Andreas and Fillard, Pierre and Thirion, Bertrand},
	year = {2010},
	note = {Series Title: Lecture Notes in Computer Science},
	pages = {200--208},
}

@article{vandermaaten_2008_VisualizingDataUsing,
	title = {Visualizing data using t-{SNE}},
	volume = {9},
	url = {http://jmlr.org/papers/v9/vandermaaten08a.html},
	number = {86},
	journal = {Journal of Machine Learning Research},
	author = {van der Maaten, Laurens and Hinton, Geoffrey},
	year = {2008},
	pages = {2579--2605},
}

@article{vanessen_2012_HumanConnectomeProject,
	title = {The {Human} {Connectome} {Project}: {A} data acquisition perspective},
	volume = {62},
	issn = {10538119},
	shorttitle = {The {Human} {Connectome} {Project}},
	url = {https://linkinghub.elsevier.com/retrieve/pii/S1053811912001954},
	doi = {10.1016/j.neuroimage.2012.02.018},
	language = {en},
	number = {4},
	urldate = {2022-08-28},
	journal = {NeuroImage},
	author = {Van Essen, D.C. and Ugurbil, K. and Auerbach, E. and Barch, D. and Behrens, T.E.J. and Bucholz, R. and Chang, A. and Chen, L. and Corbetta, M. and Curtiss, S.W. and Della Penna, S. and Feinberg, D. and Glasser, M.F. and Harel, N. and Heath, A.C. and Larson-Prior, L. and Marcus, D. and Michalareas, G. and Moeller, S. and Oostenveld, R. and Petersen, S.E. and Prior, F. and Schlaggar, B.L. and Smith, S.M. and Snyder, A.Z. and Xu, J. and Yacoub, E.},
	month = oct,
	year = {2012},
	pages = {2222--2231},
}

@article{tibshirani_1996_RegressionShrinkageSelection,
	title = {Regression {Shrinkage} and {Selection} {Via} the {Lasso}},
	volume = {58},
	copyright = {https://academic.oup.com/journals/pages/open\_access/funder\_policies/chorus/standard\_publication\_model},
	issn = {1369-7412, 1467-9868},
	url = {https://academic.oup.com/jrsssb/article/58/1/267/7027929},
	doi = {10.1111/j.2517-6161.1996.tb02080.x},
	language = {en},
	number = {1},
	urldate = {2024-08-15},
	journal = {Journal of the Royal Statistical Society Series B: Statistical Methodology},
	author = {Tibshirani, Robert},
	month = jan,
	year = {1996},
	pages = {267--288},
}

@article{tropp_2018_SimplicialFacesSet,
	title = {Simplicial {Faces} of the {Set} of {Correlation} {Matrices}},
	volume = {60},
	issn = {0179-5376, 1432-0444},
	url = {http://link.springer.com/10.1007/s00454-017-9961-0},
	doi = {10.1007/s00454-017-9961-0},
	language = {en},
	number = {2},
	urldate = {2021-09-27},
	journal = {Discrete \& Computational Geometry},
	author = {Tropp, Joel A.},
	month = sep,
	year = {2018},
	pages = {512--529},
}

@article{thanwerdas_2022_TheoreticallyComputationallyConvenient,
	title = {Theoretically and {Computationally} {Convenient} {Geometries} on {Full}-{Rank} {Correlation} {Matrices}},
	volume = {43},
	issn = {0895-4798, 1095-7162},
	url = {https://epubs.siam.org/doi/10.1137/22M1471729},
	doi = {10.1137/22M1471729},
	language = {en},
	number = {4},
	urldate = {2023-11-25},
	journal = {SIAM Journal on Matrix Analysis and Applications},
	author = {Thanwerdas, Yann and Pennec, Xavier},
	month = dec,
	year = {2022},
	pages = {1851--1872},
}

@article{thanwerdas_2021_GeodesicQuotientAffineMetrics,
	title = {Geodesic of the {Quotient}-{Affine} {Metrics} on {Full}-{Rank} {Correlation} {Matrices}},
	url = {http://arxiv.org/abs/2103.04621},
	urldate = {2021-07-15},
	journal = {arXiv:2103.04621 [math]},
	author = {Thanwerdas, Yann and Pennec, Xavier},
	month = mar,
	year = {2021},
	note = {arXiv: 2103.04621},
}

@article{simeon_2022_RiemannianGeometryFunctional,
	title = {Riemannian {Geometry} of {Functional} {Connectivity} {Matrices} for {Multi}-{Site} {Attention}-{Deficit}/{Hyperactivity} {Disorder} {Data} {Harmonization}},
	volume = {16},
	issn = {1662-5196},
	url = {https://www.frontiersin.org/articles/10.3389/fninf.2022.769274/full},
	doi = {10.3389/fninf.2022.769274},
	urldate = {2022-06-18},
	journal = {Frontiers in Neuroinformatics},
	author = {Simeon, Guillem and Piella, Gemma and Camara, Oscar and Pareto, Deborah},
	month = may,
	year = {2022},
	pages = {769274},
}

@article{siman-tov_2017_EarlyAgeRelatedFunctional,
	title = {Early {Age}-{Related} {Functional} {Connectivity} {Decline} in {High}-{Order} {Cognitive} {Networks}},
	volume = {8},
	issn = {1663-4365},
	url = {http://journal.frontiersin.org/article/10.3389/fnagi.2016.00330/full},
	doi = {10.3389/fnagi.2016.00330},
	urldate = {2021-10-04},
	journal = {Frontiers in Aging Neuroscience},
	author = {Siman-Tov, Tali and Bosak, Noam and Sprecher, Elliot and Paz, Rotem and Eran, Ayelet and Aharon-Peretz, Judith and Kahn, Itamar},
	month = jan,
	year = {2017},
}

@article{shen_2017_UsingConnectomebasedPredictive,
	title = {Using connectome-based predictive modeling to predict individual behavior from brain connectivity},
	volume = {12},
	issn = {1754-2189, 1750-2799},
	url = {https://www.nature.com/articles/nprot.2016.178},
	doi = {10.1038/nprot.2016.178},
	language = {en},
	number = {3},
	urldate = {2024-08-15},
	journal = {Nature Protocols},
	author = {Shen, Xilin and Finn, Emily S and Scheinost, Dustin and Rosenberg, Monica D and Chun, Marvin M and Papademetris, Xenophon and Constable, R Todd},
	month = mar,
	year = {2017},
	pages = {506--518},
}

@article{schoenberg_1938_MetricSpacesCompletely,
	title = {Metric {Spaces} and {Completely} {Monotone} {Functions}},
	volume = {39},
	issn = {0003486X},
	url = {https://www.jstor.org/stable/1968466?origin=crossref},
	doi = {10.2307/1968466},
	number = {4},
	urldate = {2024-09-12},
	journal = {The Annals of Mathematics},
	author = {Schoenberg, I. J.},
	month = oct,
	year = {1938},
	pages = {811},
}

@article{schaefer_2018_LocalGlobalParcellationHuman,
	title = {Local-{Global} {Parcellation} of the {Human} {Cerebral} {Cortex} from {Intrinsic} {Functional} {Connectivity} {MRI}},
	volume = {28},
	copyright = {https://academic.oup.com/journals/pages/open\_access/funder\_policies/chorus/standard\_publication\_model},
	issn = {1047-3211, 1460-2199},
	url = {https://academic.oup.com/cercor/article/28/9/3095/3978804},
	doi = {10.1093/cercor/bhx179},
	language = {en},
	number = {9},
	urldate = {2024-08-01},
	journal = {Cerebral Cortex},
	author = {Schaefer, Alexander and Kong, Ru and Gordon, Evan M and Laumann, Timothy O and Zuo, Xi-Nian and Holmes, Avram J and Eickhoff, Simon B and Yeo, B T Thomas},
	month = sep,
	year = {2018},
	pages = {3095--3114},
}

@book{scholkopf_2002_LearningKernelsSupport,
	address = {Cambridge, Mass.},
	edition = {Reprint.},
	series = {Adaptive computation and machine learning series},
	title = {Learning with kernels: support vector machines, regularization, optimization, and beyond},
	isbn = {978-0-262-53657-8 978-0-262-19475-4},
	shorttitle = {Learning with kernels},
	language = {eng},
	publisher = {MIT Press},
	author = {Schölkopf, Bernhard and Smola, Alexander Johannes},
	year = {2002},
}

@misc{schalk_2009_EEGMotorMovement,
	title = {{EEG} {Motor} {Movement}/{Imagery} {Dataset}},
	url = {https://physionet.org/content/eegmmidb/},
	doi = {10.13026/C28G6P},
	urldate = {2022-09-12},
	publisher = {physionet.org},
	author = {Schalk, Gerwin and McFarland, Dennis J and Hinterberger, Thilo and Birbaumer, Niels and Wolpaw, Jonathan R},
	year = {2009},
}

@article{schalk_2004_BCI2000GeneralPurposeBrainComputer,
	title = {{BCI2000}: {A} {General}-{Purpose} {Brain}-{Computer} {Interface} ({BCI}) {System}},
	volume = {51},
	issn = {0018-9294},
	shorttitle = {{BCI2000}},
	url = {http://ieeexplore.ieee.org/document/1300799/},
	doi = {10.1109/TBME.2004.827072},
	language = {en},
	number = {6},
	urldate = {2022-09-11},
	journal = {IEEE Transactions on Biomedical Engineering},
	author = {Schalk, G. and McFarland, D.J. and Hinterberger, T. and Birbaumer, N. and Wolpaw, J.R.},
	month = jun,
	year = {2004},
	pages = {1034--1043},
}

@article{salimi-khorshidi_2014_AutomaticDenoisingFunctional,
	title = {Automatic denoising of functional {MRI} data: {Combining} independent component analysis and hierarchical fusion of classifiers},
	volume = {90},
	issn = {10538119},
	shorttitle = {Automatic denoising of functional {MRI} data},
	url = {https://linkinghub.elsevier.com/retrieve/pii/S1053811913011956},
	doi = {10.1016/j.neuroimage.2013.11.046},
	language = {en},
	urldate = {2024-08-01},
	journal = {NeuroImage},
	author = {Salimi-Khorshidi, Gholamreza and Douaud, Gwenaëlle and Beckmann, Christian F. and Glasser, Matthew F. and Griffanti, Ludovica and Smith, Stephen M.},
	month = apr,
	year = {2014},
	pages = {449--468},
}

@book{rousseeuw_1987_RobustRegressionOutlier,
	address = {New York},
	title = {Robust regression and outlier detection},
	isbn = {978-0-471-72538-1},
	language = {eng},
	publisher = {Wiley},
	author = {Rousseeuw, Peter J. and Leroy, Annick M.},
	year = {1987},
	note = {OCLC: 219924217},
}

@article{romano_2005_ExactApproximateStepdown,
	title = {Exact and {Approximate} {Stepdown} {Methods} for {Multiple} {Hypothesis} {Testing}},
	volume = {100},
	issn = {0162-1459, 1537-274X},
	url = {http://www.tandfonline.com/doi/abs/10.1198/016214504000000539},
	doi = {10.1198/016214504000000539},
	language = {en},
	number = {469},
	urldate = {2024-06-05},
	journal = {Journal of the American Statistical Association},
	author = {Romano, Joseph P and Wolf, Michael},
	month = mar,
	year = {2005},
	pages = {94--108},
}

@article{ramdas_2017_WassersteinTwoSampleTesting,
	title = {On {Wasserstein} {Two}-{Sample} {Testing} and {Related} {Families} of {Nonparametric} {Tests}},
	volume = {19},
	issn = {1099-4300},
	url = {http://www.mdpi.com/1099-4300/19/2/47},
	doi = {10.3390/e19020047},
	language = {en},
	number = {2},
	urldate = {2021-10-02},
	journal = {Entropy},
	author = {Ramdas, Aaditya and Trillos, Nicolás and Cuturi, Marco},
	month = jan,
	year = {2017},
	pages = {47},
}

@article{pitman_1937_SignificanceTestsWhich,
	title = {Significance {Tests} {Which} {May} be {Applied} to {Samples} {From} any {Populations}},
	volume = {4},
	issn = {14666162},
	url = {https://www.jstor.org/stable/10.2307/2984124?origin=crossref},
	doi = {10.2307/2984124},
	number = {1},
	urldate = {2021-10-02},
	journal = {Supplement to the Journal of the Royal Statistical Society},
	author = {Pitman, E. J. G.},
	year = {1937},
	pages = {119},
}

@article{preti_2017_DynamicFunctionalConnectome,
	title = {The dynamic functional connectome: {State}-of-the-art and perspectives},
	volume = {160},
	issn = {10538119},
	shorttitle = {The dynamic functional connectome},
	url = {https://linkinghub.elsevier.com/retrieve/pii/S1053811916307881},
	doi = {10.1016/j.neuroimage.2016.12.061},
	language = {en},
	urldate = {2021-10-04},
	journal = {NeuroImage},
	author = {Preti, Maria Giulia and Bolton, Thomas AW and Van De Ville, Dimitri},
	month = oct,
	year = {2017},
	pages = {41--54},
}

@article{pennec_2006_IntrinsicStatisticsRiemannian,
	title = {Intrinsic {Statistics} on {Riemannian} {Manifolds}: {Basic} {Tools} for {Geometric} {Measurements}},
	volume = {25},
	issn = {0924-9907, 1573-7683},
	shorttitle = {Intrinsic {Statistics} on {Riemannian} {Manifolds}},
	url = {http://link.springer.com/10.1007/s10851-006-6228-4},
	doi = {10.1007/s10851-006-6228-4},
	language = {en},
	number = {1},
	urldate = {2021-09-27},
	journal = {Journal of Mathematical Imaging and Vision},
	author = {Pennec, Xavier},
	month = jul,
	year = {2006},
	pages = {127--154},
}

@article{onnela_2003_DynamicsMarketCorrelations,
	title = {Dynamics of market correlations: {Taxonomy} and portfolio analysis},
	volume = {68},
	issn = {1063-651X, 1095-3787},
	shorttitle = {Dynamics of market correlations},
	url = {https://link.aps.org/doi/10.1103/PhysRevE.68.056110},
	doi = {10.1103/PhysRevE.68.056110},
	language = {en},
	number = {5},
	urldate = {2022-09-12},
	journal = {Physical Review E},
	author = {Onnela, J.-P. and Chakraborti, A. and Kaski, K. and Kertész, J. and Kanto, A.},
	month = nov,
	year = {2003},
	pages = {056110},
}

@article{park_2013_StructuralFunctionalBrain,
	title = {Structural and {Functional} {Brain} {Networks}: {From} {Connections} to {Cognition}},
	volume = {342},
	issn = {0036-8075, 1095-9203},
	shorttitle = {Structural and {Functional} {Brain} {Networks}},
	url = {https://www.sciencemag.org/lookup/doi/10.1126/science.1238411},
	doi = {10.1126/science.1238411},
	language = {en},
	number = {6158},
	urldate = {2021-10-04},
	journal = {Science},
	author = {Park, H.-J. and Friston, K.},
	month = nov,
	year = {2013},
	pages = {1238411--1238411},
}

@article{park_2014_GraphIndependentComponent,
	title = {Graph {Independent} {Component} {Analysis} {Reveals} {Repertoires} of {Intrinsic} {Network} {Components} in the {Human} {Brain}},
	volume = {9},
	issn = {1932-6203},
	url = {https://dx.plos.org/10.1371/journal.pone.0082873},
	doi = {10.1371/journal.pone.0082873},
	language = {en},
	number = {1},
	urldate = {2021-10-04},
	journal = {PLoS ONE},
	author = {Park, Bumhee and Kim, Dae-Shik and Park, Hae-Jeong},
	editor = {Chialvo, Dante R.},
	month = jan,
	year = {2014},
	pages = {e82873},
}

@book{panaretos_2020_InvitationStatisticsWasserstein,
	address = {Cham},
	series = {{SpringerBriefs} in {Probability} and {Mathematical} {Statistics}},
	title = {An {Invitation} to {Statistics} in {Wasserstein} {Space}},
	isbn = {978-3-030-38437-1 978-3-030-38438-8},
	url = {http://link.springer.com/10.1007/978-3-030-38438-8},
	language = {en},
	urldate = {2021-11-06},
	publisher = {Springer International Publishing},
	author = {Panaretos, Victor M. and Zemel, Yoav},
	year = {2020},
	doi = {10.1007/978-3-030-38438-8},
}

@incollection{nielsen_2019_ClusteringHilbertProjective,
	address = {Cham},
	title = {Clustering in {Hilbert}’s {Projective} {Geometry}: {The} {Case} {Studies} of the {Probability} {Simplex} and the {Elliptope} of {Correlation} {Matrices}},
	isbn = {978-3-030-02519-9 978-3-030-02520-5},
	shorttitle = {Clustering in {Hilbert}’s {Projective} {Geometry}},
	urldate = {2021-10-04},
	booktitle = {Geometric {Structures} of {Information}},
	publisher = {Springer International Publishing},
	author = {Nielsen, Frank and Sun, Ke},
	editor = {Nielsen, Frank},
	year = {2019},
	note = {Series Title: Signals and Communication Technology},
	pages = {297--331},
}

@article{mejia_2018_ImprovedEstimationSubjectlevel,
	title = {Improved estimation of subject-level functional connectivity using full and partial correlation with empirical {Bayes} shrinkage},
	volume = {172},
	issn = {10538119},
	url = {https://linkinghub.elsevier.com/retrieve/pii/S1053811918300296},
	doi = {10.1016/j.neuroimage.2018.01.029},
	language = {en},
	urldate = {2024-08-15},
	journal = {NeuroImage},
	author = {Mejia, Amanda F. and Nebel, Mary Beth and Barber, Anita D. and Choe, Ann S. and Pekar, James J. and Caffo, Brian S. and Lindquist, Martin A.},
	month = may,
	year = {2018},
	pages = {478--491},
}

@article{mantegna_1999_HierarchicalStructureFinancial,
	title = {Hierarchical structure in financial markets},
	volume = {11},
	issn = {1434-6028},
	url = {http://link.springer.com/10.1007/s100510050929},
	doi = {10.1007/s100510050929},
	language = {en},
	number = {1},
	urldate = {2022-09-12},
	journal = {The European Physical Journal B},
	author = {Mantegna, R.N.},
	month = sep,
	year = {1999},
	pages = {193--197},
}

@inproceedings{macqueen_1967_MethodsClassificationAnalysis,
	title = {Some {Methods} for {Classification} and {Analysis} of {Multivariate} {Observations}},
	volume = {1},
	booktitle = {Proc. of the fifth berkeley symposium on mathematical statistics and probability},
	publisher = {University of California Press},
	author = {MacQueen, J. B.},
	editor = {Cam, L. M. Le and Neyman, J.},
	year = {1967},
	note = {tex.added-at: 2011-01-11T13:35:01.000+0100
tex.biburl: https://www.bibsonomy.org/bibtex/25dcdb8cd9fba78e0e791af619d61d66d/enitsirhc
tex.interhash: 8d7d4dfe7d3a06b8c9c3c2bb7aa91e28
tex.intrahash: 5dcdb8cd9fba78e0e791af619d61d66d
tex.timestamp: 2011-01-11T13:35:01.000+0100},
	pages = {281--297},
}

@article{leonardi_2013_PrincipalComponentsFunctional,
	title = {Principal components of functional connectivity: {A} new approach to study dynamic brain connectivity during rest},
	volume = {83},
	issn = {10538119},
	shorttitle = {Principal components of functional connectivity},
	url = {https://linkinghub.elsevier.com/retrieve/pii/S105381191300774X},
	doi = {10.1016/j.neuroimage.2013.07.019},
	language = {en},
	urldate = {2021-10-04},
	journal = {NeuroImage},
	author = {Leonardi, Nora and Richiardi, Jonas and Gschwind, Markus and Simioni, Samanta and Annoni, Jean-Marie and Schluep, Myriam and Vuilleumier, Patrik and Van De Ville, Dimitri},
	month = dec,
	year = {2013},
	pages = {937--950},
}

@article{ledoit_2004_WellconditionedEstimatorLargedimensional,
	title = {A well-conditioned estimator for large-dimensional covariance matrices},
	volume = {88},
	copyright = {https://www.elsevier.com/tdm/userlicense/1.0/},
	issn = {0047259X},
	url = {https://linkinghub.elsevier.com/retrieve/pii/S0047259X03000964},
	doi = {10.1016/S0047-259X(03)00096-4},
	language = {en},
	number = {2},
	urldate = {2024-05-03},
	journal = {Journal of Multivariate Analysis},
	author = {Ledoit, Olivier and Wolf, Michael},
	month = feb,
	year = {2004},
	pages = {365--411},
}

@book{lee_2012_IntroductionSmoothManifolds,
	address = {New York, NY},
	series = {Graduate {Texts} in {Mathematics}},
	title = {Introduction to {Smooth} {Manifolds}},
	volume = {218},
	isbn = {978-1-4419-9981-8 978-1-4419-9982-5},
	url = {https://link.springer.com/10.1007/978-1-4419-9982-5},
	language = {en},
	urldate = {2023-12-08},
	publisher = {Springer New York},
	author = {Lee, John M.},
	year = {2012},
	doi = {10.1007/978-1-4419-9982-5},
}

@book{lee_2018_IntroductionRiemannianManifolds,
	address = {Cham},
	edition = {2nd ed. 2018},
	series = {Graduate {Texts} in {Mathematics}},
	title = {Introduction to {Riemannian} {Manifolds}},
	isbn = {978-3-319-91755-9},
	number = {176},
	publisher = {Springer International Publishing : Imprint: Springer},
	author = {Lee, John M.},
	year = {2018},
	doi = {10.1007/978-3-319-91755-9},
}

@book{klebanov_2005_NdistancesTheirApplications,
	address = {Prague},
	edition = {1. ed},
	title = {N-distances and their applications},
	isbn = {978-80-246-1152-5},
	language = {eng},
	publisher = {Karolinum Press},
	author = {Klebanov, Lev Borisovič},
	year = {2005},
}

@book{kaufman_2005_FindingGroupsData,
	address = {Hoboken, N.J},
	series = {Wiley series in probability and mathematical statistics},
	title = {Finding groups in data: an introduction to cluster analysis},
	isbn = {978-0-471-73578-6},
	shorttitle = {Finding groups in data},
	publisher = {Wiley},
	author = {Kaufman, Leonard and Rousseeuw, Peter J.},
	year = {2005},
}

@incollection{kaufman_1990_PartitioningMedoidsProgram,
	address = {Hoboken, NJ, USA},
	title = {Partitioning {Around} {Medoids} ({Program} {PAM})},
	isbn = {978-0-470-31680-1 978-0-471-87876-6},
	url = {https://onlinelibrary.wiley.com/doi/10.1002/9780470316801.ch2},
	language = {en},
	urldate = {2021-10-02},
	booktitle = {Wiley {Series} in {Probability} and {Statistics}},
	publisher = {John Wiley \& Sons, Inc.},
	author = {Kaufman, Leonard and Rousseeuw, Peter J.},
	month = mar,
	year = {1990},
	doi = {10.1002/9780470316801.ch2},
	pages = {68--125},
}

@article{jayasumana_2015_KernelMethodsRiemannian,
	title = {Kernel {Methods} on {Riemannian} {Manifolds} with {Gaussian} {RBF} {Kernels}},
	volume = {37},
	copyright = {https://ieeexplore.ieee.org/Xplorehelp/downloads/license-information/IEEE.html},
	issn = {0162-8828, 2160-9292},
	url = {http://ieeexplore.ieee.org/document/7063231/},
	doi = {10.1109/TPAMI.2015.2414422},
	number = {12},
	urldate = {2024-09-12},
	journal = {IEEE Transactions on Pattern Analysis and Machine Intelligence},
	author = {Jayasumana, Sadeep and Hartley, Richard and Salzmann, Mathieu and Li, Hongdong and Harandi, Mehrtash},
	month = dec,
	year = {2015},
	pages = {2464--2477},
}

@book{huber_1981_RobustStatistics,
	address = {New York},
	series = {Wiley series in probability and mathematical statistics},
	title = {Robust statistics},
	isbn = {978-0-471-41805-4},
	publisher = {Wiley},
	author = {Huber, Peter J.},
	year = {1981},
}

@article{honnorat_2024_RiemannianFrameworksHarmonization,
	title = {Riemannian frameworks for the harmonization of resting-state functional {MRI} scans},
	volume = {91},
	issn = {13618415},
	url = {https://linkinghub.elsevier.com/retrieve/pii/S1361841523003031},
	doi = {10.1016/j.media.2023.103043},
	language = {en},
	urldate = {2024-09-13},
	journal = {Medical Image Analysis},
	author = {Honnorat, Nicolas and Seshadri, Sudha and Killiany, Ron and Blangero, John and Glahn, David C. and Fox, Peter and Habes, Mohamad},
	month = jan,
	year = {2024},
	pages = {103043},
}

@book{hall_2015_LieGroupsLie,
	address = {Cham ; New York},
	edition = {Second edition},
	series = {Graduate texts in mathematics},
	title = {Lie groups, {Lie} algebras, and representations: an elementary introduction},
	isbn = {978-3-319-13466-6},
	shorttitle = {Lie groups, {Lie} algebras, and representations},
	number = {222},
	publisher = {Springer},
	author = {Hall, Brian C.},
	year = {2015},
	note = {OCLC: ocn910324548},
}

@article{griffanti_2014_ICAbasedArtefactRemoval,
	title = {{ICA}-based artefact removal and accelerated {fMRI} acquisition for improved resting state network imaging},
	volume = {95},
	issn = {10538119},
	url = {https://linkinghub.elsevier.com/retrieve/pii/S1053811914001815},
	doi = {10.1016/j.neuroimage.2014.03.034},
	language = {en},
	urldate = {2024-08-01},
	journal = {NeuroImage},
	author = {Griffanti, Ludovica and Salimi-Khorshidi, Gholamreza and Beckmann, Christian F. and Auerbach, Edward J. and Douaud, Gwenaëlle and Sexton, Claire E. and Zsoldos, Enikő and Ebmeier, Klaus P. and Filippini, Nicola and Mackay, Clare E. and Moeller, Steen and Xu, Junqian and Yacoub, Essa and Baselli, Giuseppe and Ugurbil, Kamil and Miller, Karla L. and Smith, Stephen M.},
	month = jul,
	year = {2014},
	pages = {232--247},
}

@article{grubisic_2007_EfficientRankReduction,
	title = {Efficient rank reduction of correlation matrices},
	volume = {422},
	issn = {00243795},
	url = {https://linkinghub.elsevier.com/retrieve/pii/S0024379506005222},
	doi = {10.1016/j.laa.2006.11.024},
	language = {en},
	number = {2-3},
	urldate = {2021-10-04},
	journal = {Linear Algebra and its Applications},
	author = {Grubišić, Igor and Pietersz, Raoul},
	month = apr,
	year = {2007},
	pages = {629--653},
}

@article{glasser_2013_MinimalPreprocessingPipelines,
	title = {The minimal preprocessing pipelines for the {Human} {Connectome} {Project}},
	volume = {80},
	issn = {10538119},
	url = {https://linkinghub.elsevier.com/retrieve/pii/S1053811913005053},
	doi = {10.1016/j.neuroimage.2013.04.127},
	language = {en},
	urldate = {2024-08-01},
	journal = {NeuroImage},
	author = {Glasser, Matthew F. and Sotiropoulos, Stamatios N. and Wilson, J. Anthony and Coalson, Timothy S. and Fischl, Bruce and Andersson, Jesper L. and Xu, Junqian and Jbabdi, Saad and Webster, Matthew and Polimeni, Jonathan R. and Van Essen, David C. and Jenkinson, Mark},
	month = oct,
	year = {2013},
	pages = {105--124},
}

@article{goldberger_2000_PhysioBankPhysioToolkitPhysioNet,
	title = {{PhysioBank}, {PhysioToolkit}, and {PhysioNet}: {Components} of a {New} {Research} {Resource} for {Complex} {Physiologic} {Signals}},
	volume = {101},
	issn = {0009-7322, 1524-4539},
	shorttitle = {{PhysioBank}, {PhysioToolkit}, and {PhysioNet}},
	url = {https://www.ahajournals.org/doi/10.1161/01.CIR.101.23.e215},
	doi = {10.1161/01.CIR.101.23.e215},
	language = {en},
	number = {23},
	urldate = {2022-09-11},
	journal = {Circulation},
	author = {Goldberger, Ary L. and Amaral, Luis A. N. and Glass, Leon and Hausdorff, Jeffrey M. and Ivanov, Plamen Ch. and Mark, Roger G. and Mietus, Joseph E. and Moody, George B. and Peng, Chung-Kang and Stanley, H. Eugene},
	month = jun,
	year = {2000},
}

@article{ginestet_2017_HypothesisTestingNetwork,
	title = {Hypothesis testing for network data in functional neuroimaging},
	volume = {11},
	issn = {1932-6157},
	url = {https://projecteuclid.org/journals/annals-of-applied-statistics/volume-11/issue-2/Hypothesis-testing-for-network-data-in-functional-neuroimaging/10.1214/16-AOAS1015.full},
	doi = {10.1214/16-AOAS1015},
	number = {2},
	urldate = {2021-10-04},
	journal = {The Annals of Applied Statistics},
	author = {Ginestet, Cedric E. and Li, Jun and Balachandran, Prakash and Rosenberg, Steven and Kolaczyk, Eric D.},
	month = jun,
	year = {2017},
}

@article{fletcher_2009_GeometricMedianRiemannian,
	title = {The geometric median on {Riemannian} manifolds with application to robust atlas estimation},
	volume = {45},
	issn = {10538119},
	url = {https://linkinghub.elsevier.com/retrieve/pii/S1053811908012019},
	doi = {10.1016/j.neuroimage.2008.10.052},
	language = {en},
	number = {1},
	urldate = {2021-10-02},
	journal = {NeuroImage},
	author = {Fletcher, P. Thomas and Venkatasubramanian, Suresh and Joshi, Sarang},
	month = mar,
	year = {2009},
	pages = {S143--S152},
}

@article{fletcher_2004_PrincipalGeodesicAnalysis,
	title = {Principal {Geodesic} {Analysis} for the {Study} of {Nonlinear} {Statistics} of {Shape}},
	volume = {23},
	issn = {0278-0062},
	url = {http://ieeexplore.ieee.org/document/1318725/},
	doi = {10.1109/TMI.2004.831793},
	language = {en},
	number = {8},
	urldate = {2021-10-03},
	journal = {IEEE Transactions on Medical Imaging},
	author = {Fletcher, P.T. and Lu, C. and Pizer, S.M. and Joshi, S.},
	month = aug,
	year = {2004},
	pages = {995--1005},
}

@article{finn_2015_FunctionalConnectomeFingerprinting,
	title = {Functional connectome fingerprinting: identifying individuals using patterns of brain connectivity},
	volume = {18},
	issn = {1097-6256, 1546-1726},
	shorttitle = {Functional connectome fingerprinting},
	url = {https://www.nature.com/articles/nn.4135},
	doi = {10.1038/nn.4135},
	language = {en},
	number = {11},
	urldate = {2021-10-04},
	journal = {Nature Neuroscience},
	author = {Finn, Emily S and Shen, Xilin and Scheinost, Dustin and Rosenberg, Monica D and Huang, Jessica and Chun, Marvin M and Papademetris, Xenophon and Constable, R Todd},
	month = nov,
	year = {2015},
	pages = {1664--1671},
}

@inproceedings{feragen_2015_GeodesicExponentialKernels,
	address = {Boston, MA, USA},
	title = {Geodesic exponential kernels: {When} curvature and linearity conflict},
	isbn = {978-1-4673-6964-0},
	shorttitle = {Geodesic exponential kernels},
	url = {http://ieeexplore.ieee.org/document/7298922/},
	doi = {10.1109/CVPR.2015.7298922},
	urldate = {2022-07-26},
	booktitle = {2015 {IEEE} {Conference} on {Computer} {Vision} and {Pattern} {Recognition} ({CVPR})},
	publisher = {IEEE},
	author = {Feragen, Aasa and Lauze, Francois and Hauberg, Soren},
	month = jun,
	year = {2015},
	pages = {3032--3042},
}

@article{engel_2012_SurveyDimensionReduction,
	title = {A {Survey} of {Dimension} {Reduction} {Methods} for {High}-dimensional {Data} {Analysis} and {Visualization}},
	url = {http://drops.dagstuhl.de/opus/volltexte/2012/3747/},
	doi = {10.4230/OASICS.VLUDS.2011.135},
	language = {en},
	urldate = {2021-11-17},
	author = {Engel, Daniel and Hüttenberger, Lars and Hamann, Bernd},
	collaborator = {Herbstritt, Marc},
	year = {2012},
	note = {Artwork Size: 15 pages
Medium: application/pdf
Publisher: Schloss Dagstuhl - Leibniz-Zentrum fuer Informatik GmbH, Wadern/Saarbruecken, Germany},
	pages = {15 pages},
}

@article{dosenbach_2010_PredictionIndividualBrain,
	title = {Prediction of {Individual} {Brain} {Maturity} {Using} {fMRI}},
	volume = {329},
	issn = {0036-8075, 1095-9203},
	url = {https://www.sciencemag.org/lookup/doi/10.1126/science.1194144},
	doi = {10.1126/science.1194144},
	language = {en},
	number = {5997},
	urldate = {2021-10-04},
	journal = {Science},
	author = {Dosenbach, Nico U. F. and Nardos, Binyam and Cohen, Alexander L. and Fair, Damien A. and Power, Jonathan D. and Church, Jessica A. and Nelson, Steven M. and Wig, Gagan S. and Vogel, Alecia C. and Lessov-Schlaggar, Christina N. and Barnes, Kelly Anne and Dubis, Joseph W. and Feczko, Eric and Coalson, Rebecca S. and Pruett, John R. and Barch, Deanna M. and Petersen, Steven E. and Schlaggar, Bradley L.},
	month = sep,
	year = {2010},
	pages = {1358--1361},
}

@book{demmel_1997_AppliedNumericalLinear,
	address = {Philadelphia},
	title = {Applied numerical linear algebra},
	isbn = {978-0-89871-389-3},
	publisher = {Society for Industrial and Applied Mathematics},
	author = {Demmel, James W.},
	year = {1997},
}

@phdthesis{david_2019_RiemannianQuotientStructure,
	type = {{PhD} {Thesis}},
	title = {A {Riemannian} {Quotient} {Structure} for {Correlation} {Matrices} with {Applications} to {Data} {Science}},
	school = {Claremont Graduate University},
	author = {David, Paul},
	year = {2019},
}

@book{cohen_2014_AnalyzingNeuralTime,
	address = {Cambridge, Massachusetts},
	series = {Issues in clinical and cognitive neuropsychology},
	title = {Analyzing neural time series data: theory and practice},
	isbn = {978-0-262-01987-3},
	shorttitle = {Analyzing neural time series data},
	publisher = {The MIT Press},
	author = {Cohen, Mike X.},
	year = {2014},
}

@article{chen_2010_ShrinkageAlgorithmsMMSE,
	title = {Shrinkage {Algorithms} for {MMSE} {Covariance} {Estimation}},
	volume = {58},
	issn = {1053-587X, 1941-0476},
	url = {http://ieeexplore.ieee.org/document/5484583/},
	doi = {10.1109/TSP.2010.2053029},
	number = {10},
	urldate = {2022-07-26},
	journal = {IEEE Transactions on Signal Processing},
	author = {Chen, Yilun and Wiesel, Ami and Eldar, Yonina C. and Hero, Alfred O.},
	month = oct,
	year = {2010},
	pages = {5016--5029},
}

@book{carmo_1992_RiemannianGeometry,
	address = {Boston},
	series = {Mathematics. {Theory} \& applications},
	title = {Riemannian geometry},
	isbn = {978-0-8176-3490-2 978-3-7643-3490-1},
	language = {eng},
	publisher = {Birkhäuser},
	author = {Carmo, Manfredo Perdigão do},
	year = {1992},
}

@article{calinski_1974_DendriteMethodCluster,
	title = {A dendrite method for cluster analysis},
	volume = {3},
	issn = {0361-0926},
	url = {http://www.tandfonline.com/doi/abs/10.1080/03610927408827101},
	doi = {10.1080/03610927408827101},
	language = {en},
	number = {1},
	urldate = {2021-10-02},
	journal = {Communications in Statistics - Theory and Methods},
	author = {Calinski, T. and Harabasz, J.},
	year = {1974},
	pages = {1--27},
}

@book{buhlmann_2011_StatisticsHighDimensionalData,
	address = {Berlin, Heidelberg},
	series = {Springer {Series} in {Statistics}},
	title = {Statistics for {High}-{Dimensional} {Data}: {Methods}, {Theory} and {Applications}},
	copyright = {https://www.springernature.com/gp/researchers/text-and-data-mining},
	isbn = {978-3-642-20191-2 978-3-642-20192-9},
	shorttitle = {Statistics for {High}-{Dimensional} {Data}},
	url = {https://link.springer.com/10.1007/978-3-642-20192-9},
	language = {en},
	urldate = {2024-07-24},
	publisher = {Springer Berlin Heidelberg},
	author = {Bühlmann, Peter and Van De Geer, Sara},
	year = {2011},
	doi = {10.1007/978-3-642-20192-9},
}

@article{brookes_2011_MeasuringFunctionalConnectivity,
	title = {Measuring functional connectivity using {MEG}: {Methodology} and comparison with {fcMRI}},
	volume = {56},
	issn = {10538119},
	shorttitle = {Measuring functional connectivity using {MEG}},
	url = {https://linkinghub.elsevier.com/retrieve/pii/S1053811911002102},
	doi = {10.1016/j.neuroimage.2011.02.054},
	language = {en},
	number = {3},
	urldate = {2022-09-12},
	journal = {NeuroImage},
	author = {Brookes, Matthew J. and Hale, Joanne R. and Zumer, Johanna M. and Stevenson, Claire M. and Francis, Susan T. and Barnes, Gareth R. and Owen, Julia P. and Morris, Peter G. and Nagarajan, Srikantan S.},
	month = jun,
	year = {2011},
	pages = {1082--1104},
}

@book{boumal_2023_IntroductionOptimizationSmooth,
	edition = {1},
	title = {An {Introduction} to {Optimization} on {Smooth} {Manifolds}},
	copyright = {https://www.cambridge.org/core/terms},
	isbn = {978-1-00-916616-4 978-1-00-916617-1 978-1-00-916615-7},
	url = {https://www.cambridge.org/core/product/identifier/9781009166164/type/book},
	urldate = {2024-06-07},
	publisher = {Cambridge University Press},
	author = {Boumal, Nicolas},
	month = mar,
	year = {2023},
	doi = {10.1017/9781009166164},
}

@book{borg_1997_ModernMultidimensionalScaling,
	address = {New York},
	series = {Springer series in statistics},
	title = {Modern multidimensional scaling: theory and applications},
	isbn = {978-0-387-94845-4},
	shorttitle = {Modern multidimensional scaling},
	publisher = {Springer},
	author = {Borg, Ingwer and Groenen, Patrick J. F.},
	year = {1997},
}

@article{bonanno_2003_TopologyCorrelationbasedMinimal,
	title = {Topology of correlation-based minimal spanning trees in real and model markets},
	volume = {68},
	issn = {1063-651X, 1095-3787},
	url = {https://link.aps.org/doi/10.1103/PhysRevE.68.046130},
	doi = {10.1103/PhysRevE.68.046130},
	language = {en},
	number = {4},
	urldate = {2022-09-12},
	journal = {Physical Review E},
	author = {Bonanno, Giovanni and Caldarelli, Guido and Lillo, Fabrizio and Mantegna, Rosario N.},
	month = oct,
	year = {2003},
	pages = {046130},
}

@article{biswas_2014_NonparametricTwosampleTest,
	title = {A nonparametric two-sample test applicable to high dimensional data},
	volume = {123},
	issn = {0047259X},
	url = {https://linkinghub.elsevier.com/retrieve/pii/S0047259X13001966},
	doi = {10.1016/j.jmva.2013.09.004},
	language = {en},
	urldate = {2021-10-02},
	journal = {Journal of Multivariate Analysis},
	author = {Biswas, Munmun and Ghosh, Anil K.},
	month = jan,
	year = {2014},
	pages = {160--171},
}

@article{biswal_1995_FunctionalConnectivityMotor,
	title = {Functional connectivity in the motor cortex of resting human brain using echo-planar mri},
	volume = {34},
	issn = {07403194, 15222594},
	url = {https://onlinelibrary.wiley.com/doi/10.1002/mrm.1910340409},
	doi = {10.1002/mrm.1910340409},
	language = {en},
	number = {4},
	urldate = {2021-10-04},
	journal = {Magnetic Resonance in Medicine},
	author = {Biswal, Bharat and Zerrin Yetkin, F. and Haughton, Victor M. and Hyde, James S.},
	month = oct,
	year = {1995},
	pages = {537--541},
}

@article{bilker_2012_DevelopmentAbbreviatedNineItem,
	title = {Development of {Abbreviated} {Nine}-{Item} {Forms} of the {Raven}’s {Standard} {Progressive} {Matrices} {Test}},
	volume = {19},
	issn = {1073-1911, 1552-3489},
	url = {http://journals.sagepub.com/doi/10.1177/1073191112446655},
	doi = {10.1177/1073191112446655},
	language = {en},
	number = {3},
	urldate = {2024-08-14},
	journal = {Assessment},
	author = {Bilker, Warren B. and Hansen, John A. and Brensinger, Colleen M. and Richard, Jan and Gur, Raquel E. and Gur, Ruben C.},
	month = sep,
	year = {2012},
	pages = {354--369},
}

\end{document}